\documentclass[pdflatex,sn-mathphys-num,iicol]{sn-jnl}

\usepackage{amsfonts}       

\usepackage{manyfoot}%
\usepackage{graphicx}
\usepackage{caption}
\usepackage{subcaption}
\usepackage{amsmath, amssymb}
\usepackage[title]{appendix}

\usepackage{xcolor}
\newcommand{\x}[1]{{\color{black}{#1}}}

\usepackage{geometry}
\begin{document}

\title[Article Title]{Multi-scale Topology Optimization using Neural Networks}

\author[1]{\fnm{Hongrui}\sur{Chen}}  
\author[2]{\fnm{Xingchen}\sur{Liu}}  
\author*[1]{\fnm{Levent Burak}\sur{Kara}}\email{lkara@cmu.edu} 

\affil[1]{%
  \orgname{Carnegie Mellon University}
}

\affil[2]{%
  \orgname{Intact Solutions}
}



\abstract{
A long-standing challenge in multi-scale structural design is ensuring proper connectivity between the constituent cells as each cell is being optimized toward its theoretical performance limit. We propose a new method for multi-scale topology optimization that seamlessly ensures compatibility across neighboring microstructure cells without the need for a catalog of unit cells and associated connectivity rules. Our approach consists of a topology neural network that encodes each cell's microstructure as well as the distribution of the optimized cells within the design domain as a continuous and differentiable density field. Each cell is optimized based on a prescribed elasticity tensor that also accommodates in-plane rotations. The neural network takes as input the local coordinates within a cell to represent the density distribution within a cell, as well as the global coordinates of each cell to generate spatially varying microstructure cells. As such, our approach models an n-dimensional multi-scale optimization problem as a 2n-dimensional inverse homogenization problem using neural networks. During the inverse homogenization of each unit cell, we extend the boundary of each cell by scaling the input coordinates such that the boundaries of neighboring cells are blended and thus jointly optimized. This enables inverse homogenization of the cells to be consistently connected without introducing discontinuities at the cell boundaries. We demonstrate our method through the design and optimization of several graded multi-scale structures.    }

\maketitle

\section{INTRODUCTION}

There has been a recent increase in machine learning-driven topology and multi-scale structures optimization approaches, particularly using neural networks for performing structural optimization. Both data-driven \cite{nie2021topologygan,maze2023diffusion,nobari2024nito,li2019non,oh2019deep} and direct training-based \cite{Chandrasekhar2021,Chandrasekhar2021MM,Chandrasekhar2021Fourier,chen2023cond,sridhara2022generalized,zehnder2021ntopo} approaches have been explored. Data-driven approaches typically require large training sets and long training times. They perform instant optimal topology generation during inference time. Direct training approaches use the neural network to represent the density field of a single or a group of designs. Prior works on direct training approaches focus on using the neural network with spatial coordinates as input and density as the output to represent the geometry \cite{chen2023cond, sridhara2022generalized, Chandrasekhar2021}. Inverse homogenization refers to the optimization of microstructure with the desired property \cite{sigmund1994materials,wu2021topology}. Sridhara et al. \cite{sridhara2022generalized} applied direct neural network-based inverse homogenization in designing single microstructure cells. Unfortunately, no direct neural network-based optimization for multi-scale structures has been developed. One key challenge from microstructure design to multi-scale structure is ensuring good connectivity between cells \cite{wu2021topology}.  Connecting individual optimized microstructure cells into a multi-scale structure is called de-homogenization \cite{wu2021topology}. Cell rotation is included in de-homogenization in which cells can perform very close to the theoretical limit\cite{wu2021topology}. De-homogenization is a post-processing step to obtain a well-connected multi-scale design introduced by Pantz and Trabelsi \cite{pantz2008post}. During inverse homogenization, connectivity can also be encouraged where Garner et al. \cite{garner2019compatibility} combined adjacent cells and Liu et al. \cite{liu2020two} formulated a constraint on the boundary of each cell. However, these previous methods did not incorporate rotation during inverse homogenization. On the other hand, de-homogenization relies on post-processing to add rotation. \footnote{Example code is available at \url{https://github.com/HongRayChen/MSNN}}

In this work, we aim to use a neural network to represent the design of a multi-scale structure such that connectivity and rotation can be enforced. We configure a neural network to take in both local and global spatial coordinates to represent multi-scale structural designs. We focus on 2D multi-scale structures; hence, the coordinates are two-dimensional. The local coordinates represent the micro-scale microstructure. The global coordinates correspond with the macro-scale cell location. Given the local and global coordinate pairs as input, the neural network outputs the density for that specific location as a value between 0 and 1. During training\footnote{Here ``training'' refers to optimization and back-propagation. Training has been used in previous related works \cite{Chandrasekhar2021,zehnder2021ntopo} but this is different from data-driven machine learning}, the microstructure for each cell is optimized to maximize or minimize a specific material property, derived from the elasticity tensor. We extend the unit cell boundary to its neighboring cells during inverse homogenization to encourage connectivity between cells. Since the inputs to the neural network are spatial coordinates, transformation matrices, for example, scaling, and rotation, can be applied to the coordinates. We apply rotation matrices to the local input coordinates such that the cell is oriented similarly to de-homogenization\cite{pantz2008post}. The difference is that in our approach the orientation of the cell is co-optimized during the training of the neural network instead of a post-processing step. 

In the training of a conventional machine learning model, a loss function is used to back-propagate and update the parameters in the model. Similarly, we also formulate a combined loss function for the optimization of multi-scale structures. The combined loss includes the target volume fraction for each cell, the error w.r.t target material property, and an L1 loss between cell boundaries to encourage microstructure connectivity. Since the inputs to the neural network are local and global coordinates, one can sample the coordinates at any point in the domain without interpolation. During the optimization, we sample the coordinates at regular intervals which corresponds to the square element used in Finite Element Analysis (FEA). The FEA module operates separately from the machine learning pipeline but outputs the objective value during forward propagation and sensitivities during backpropagation. During the training of the neural network, the network gradually learns the local geometry of each microstructure cell. Once training is finished, we can sample the coordinates at the same or finer interval such that a higher resolution result \x{with smooth and crisp boundary} can be obtained. 

Our main contributions are: 
\begin{itemize}
    \item a neural network-based optimization of multi-scale structure
    \item a method to enforce connectivity for multi-scale design 
    \item accommodates rotation of cells with neural network geometry representation 
\end{itemize}

\section{RELATED WORK}
\textit{Multi-scale topology optimization}:
Multi-scale structures offer the potential to deliver outstanding performance, being inherently lightweight, durable, and versatile. Bends{\o}e and Kikuchi \cite{bens0e1988generating} introduced the homogenization approach for topology optimization. The SIMP method \cite{bendsoe1989optimal,zhou1991coc}  considers the relative material density in each element of the Finite Element (FE) mesh as design variables, allowing for a simpler interpretation and optimized designs with more clearly defined features. A comprehensive review by Wu et al. \cite{wu2021topology} discusses the existing approaches of multi-scale topology optimization and analysis of their strength and applications. Research that targets improving the compatibility of multi-scale structures includes Garner et al. \cite{garner2019compatibility}, where adjacent cells were combined when inverse homogenization is performed such that compatibility can be improved. However, adding neighboring cells to the optimization will increase the computation time for FE analysis. We instead extend the cell boundary into neighboring cells where the FE resolution remains the same. Liu et al. \cite{liu2020two} formulated a constraint on the boundary of each cell such that given n types of desired microstructure, all these n types share the same boundary. This allows the optimized cell designs to be mutually compatible. One possible limitation is that the total types of cell designs are limited and as the number of cell designs increases, the restriction on the boundary can be increased. In this work, we establish connectivity between neighboring cells, where the limitation on geometry is less. Connectivity is also important when designing multi-scale structures given a material exemplar. Senhora et al. \cite{senhora2022optimally} use spinodal architected materials to place material along principal stress directions. A library of microstructure can be pre-generated and connectivity between cells can be improved through mapping \cite{zhu2017two} or inherent connectivity of the microstructure dataset \cite{vijayakumaran2025consistent}. However, the microstructure performance gamut can be limited by the generated dataset.


\textit{Direct topology optimization with neural networks}:
We refer to direct topology optimization methods as those that do not use any prior data but rather train a neural network in a self-supervised manner for learning the optimal density distribution/topology. Chandrasekhar and Suresh \cite{Chandrasekhar2021} explored a direct approach where the density field is parameterized using a neural network. The direct approach has also been extended to designing microstructure cells \cite{sridhara2022generalized}. We further extend upon single-cell design to formulate a neural network for graded multi-scale structure design. 

Fourier projection-based neural network for length scale control \cite{Chandrasekhar2021Fourier} and application for multi-material topology optimization \cite{Chandrasekhar2021MM} has also been explored. Deng and To \cite{deng2020topology} propose topology optimization with Deep Representation Learning, with a similar concept of re-parametrization, and demonstrate the effectiveness of the proposed method on minimum compliance and stress-constrained problems. Deng and To \cite{deng2021parametric} also propose a neural network-based method for level-set topology optimization, where a fully connected deep neural network describes the implicit function of the level-set. Zehnder et al. \cite{zehnder2021ntopo} effectively leverage neural representations in the context of mesh-free topology optimization and use multilayer perceptrons to parameterize both density and displacement fields. It enables self-supervised learning of continuous solution spaces for topology optimization problems. Mai et al. \cite{mai2023physics} develop a similar approach for the optimum design of truss structures. Hoyer et al. \cite{hoyer2019neural} use CNNs for density parametrization and directly enforce the constraints in each iteration, reducing the loss function to compliance only. They observe that the CNN solutions are qualitatively different from the baselines and often involve simpler and more effective structures. Zhang at al. \cite{zhang2021tonr} adopt a similar strategy and show solutions for different optimization problems including stress-constrained problems and compliant mechanism design.

\begin{figure*}
\centering

\includegraphics[width=\textwidth]{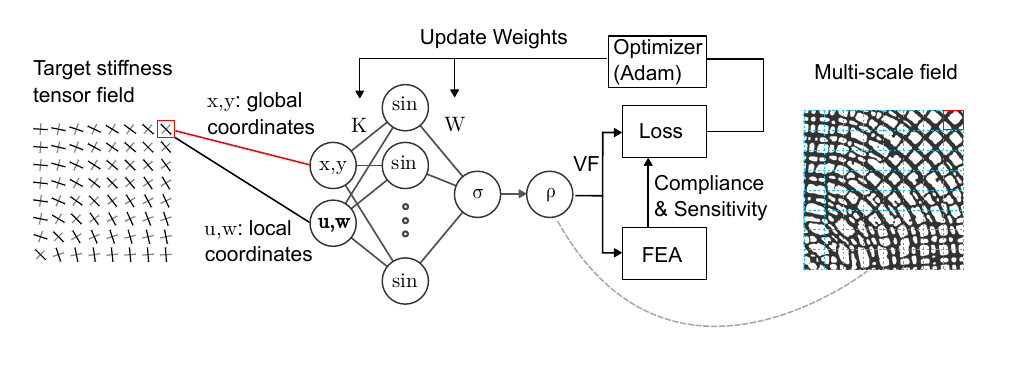}

\caption{ The neural network outputs density $\rho$ at each coordinate point. By sampling coordinate points across the design domain, we obtain the density field. From the density field, we calculate the current volume fraction and the homogenized stiffness tensor from an FEA solver. The homogenized stiffness tensor and volume fraction are then formulated as a loss function which is used in backpropagation of the training process until convergence. }

\label{fig:flowchart}
\end{figure*}
\begin{figure*}

\centering

\begin{subfigure}[t]{0.45\textwidth}
\centering
\includegraphics[width=\textwidth]{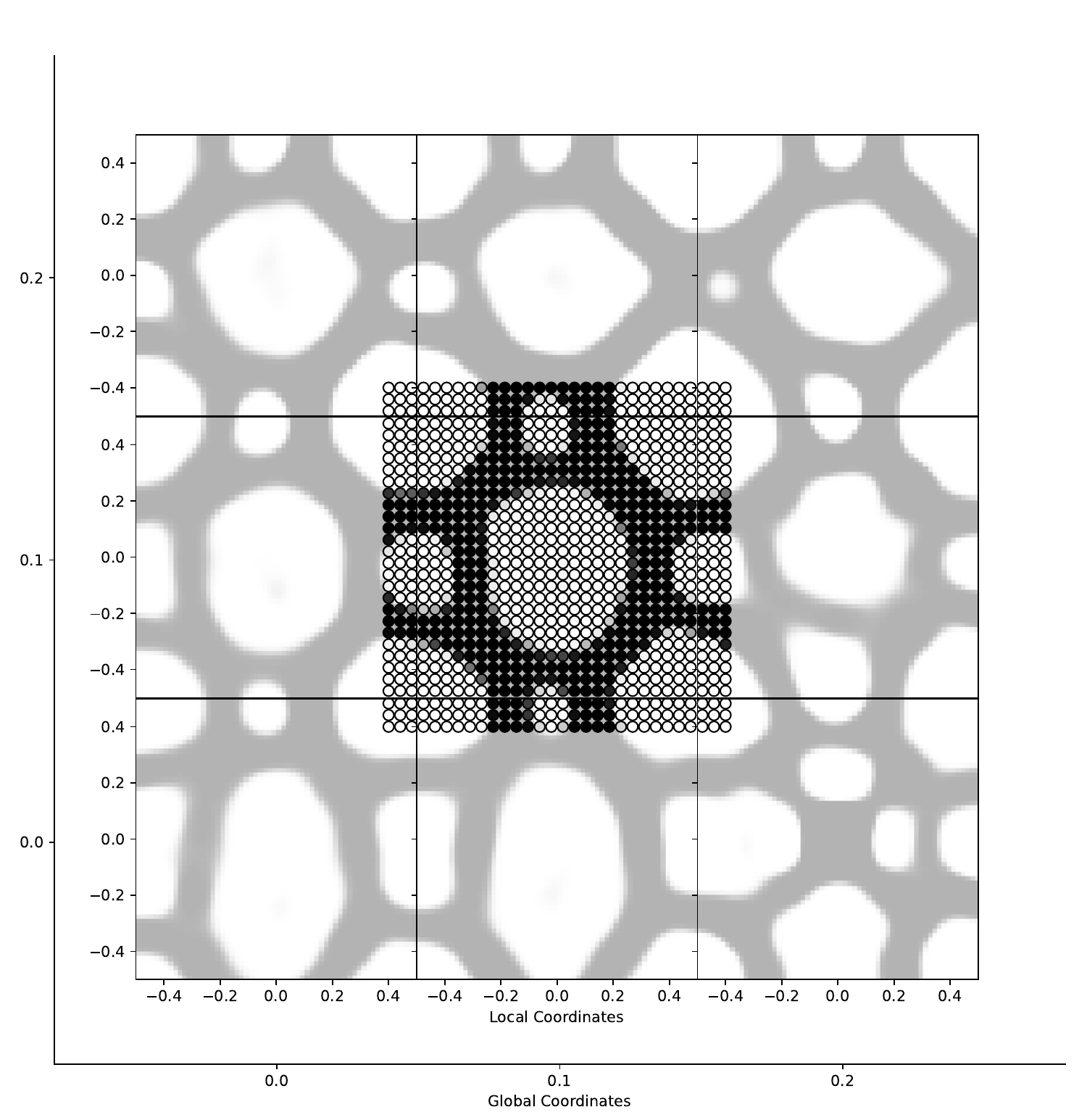}
\caption{$0^\circ$ rotation }
\end{subfigure}
\qquad
\begin{subfigure}[t]{0.45\textwidth}
\centering
\includegraphics[width=\textwidth]{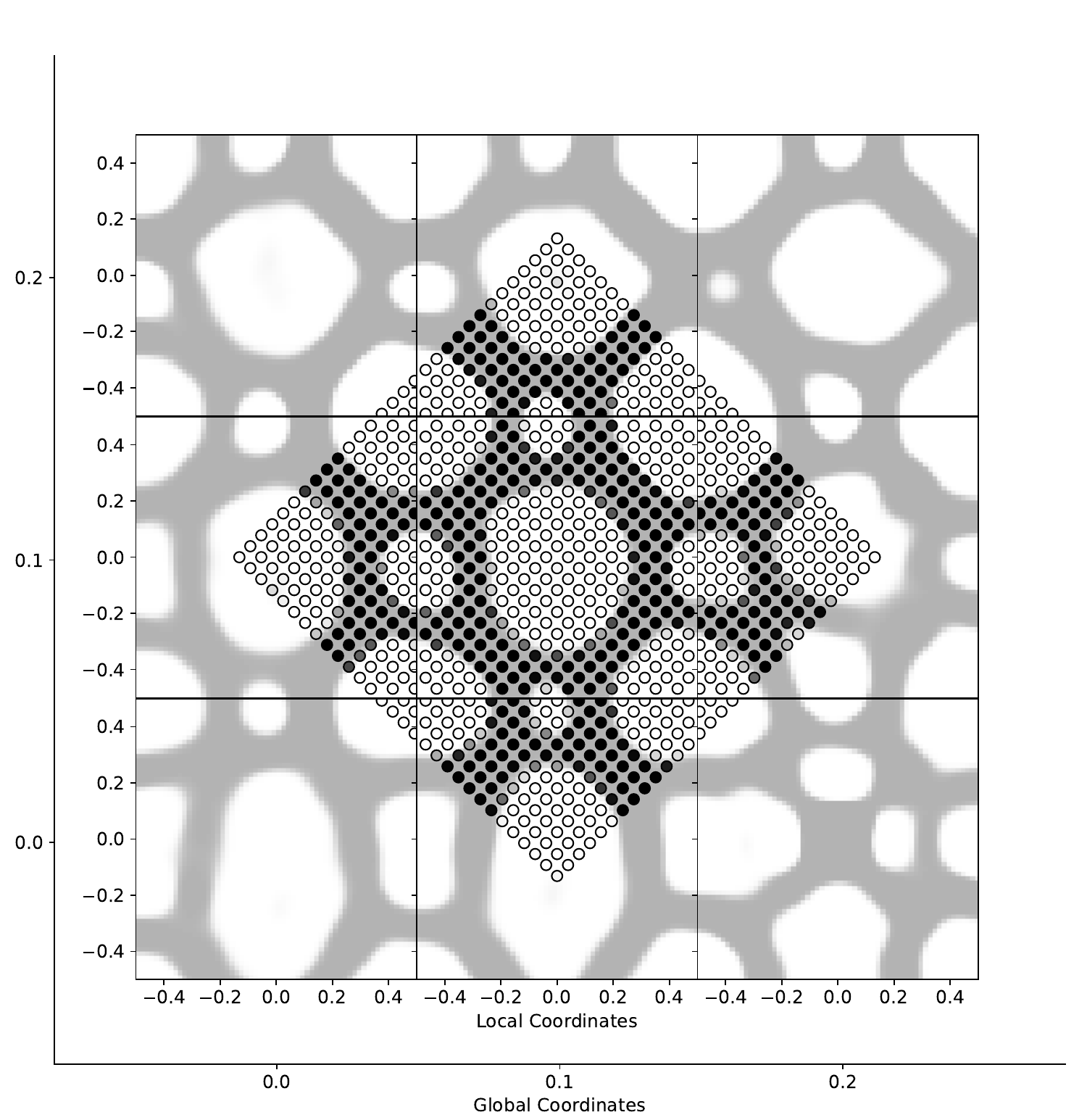}
\caption{$45^\circ$ rotation}
\end{subfigure}

\caption{Sampling inside the continuous topology domain for finite element analysis. When the rotation is at $0^\circ$, the boundary is extended by 1.2 times to ensure all 8 neighboring cells are covered. With a $45^\circ$ rotation, a 1.6 times increase is applied. }

\label{fig:febc_extension}
\end{figure*}
\begin{figure*}

\centering

\begin{subfigure}[t]{0.6\textwidth}
\centering
\includegraphics[width=\textwidth]{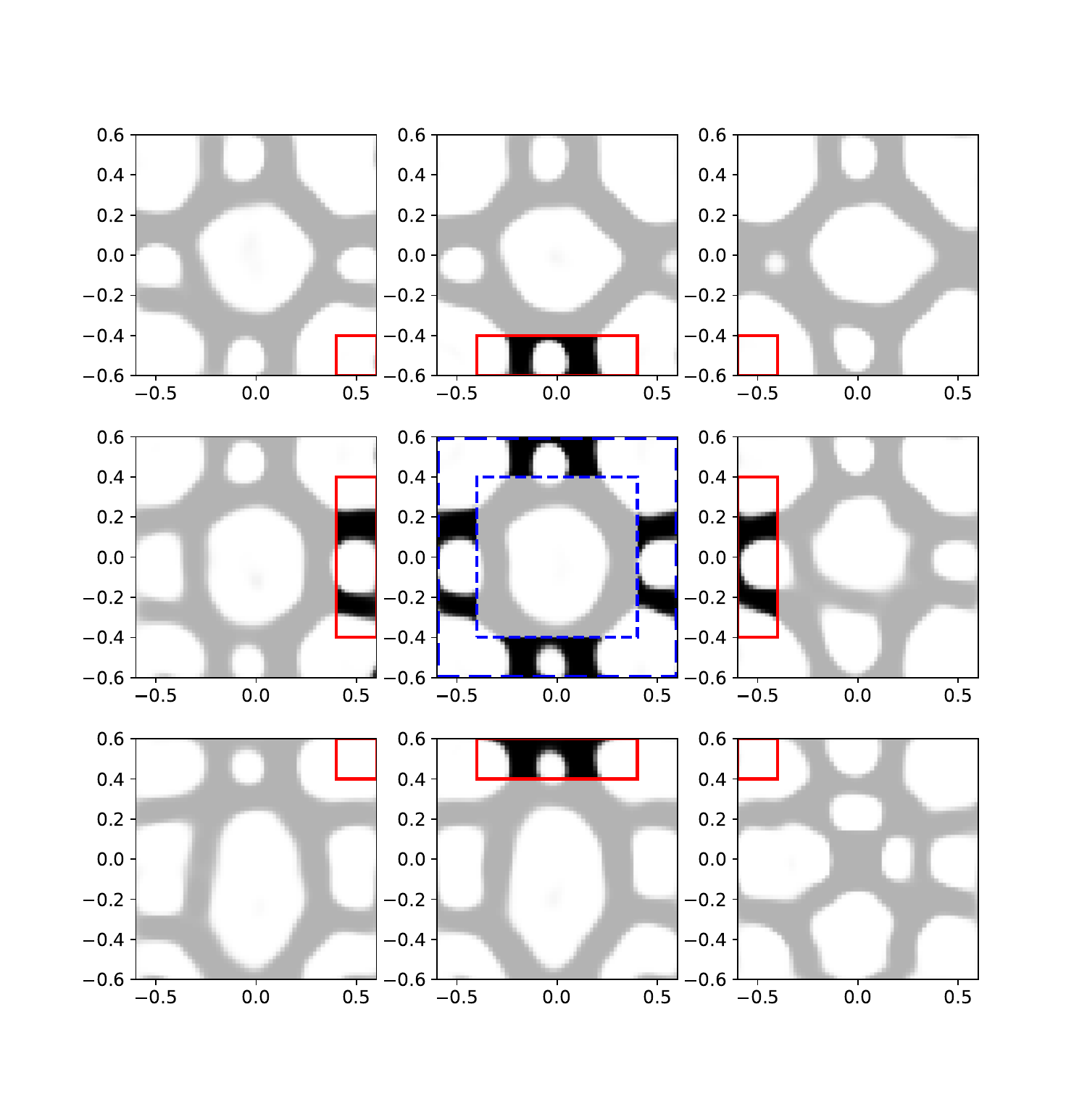}
\caption{Graphical illustration of boundary loss regions}
\end{subfigure}

\begin{subfigure}[t]{0.3\textwidth}
\centering
\includegraphics[width=\textwidth]{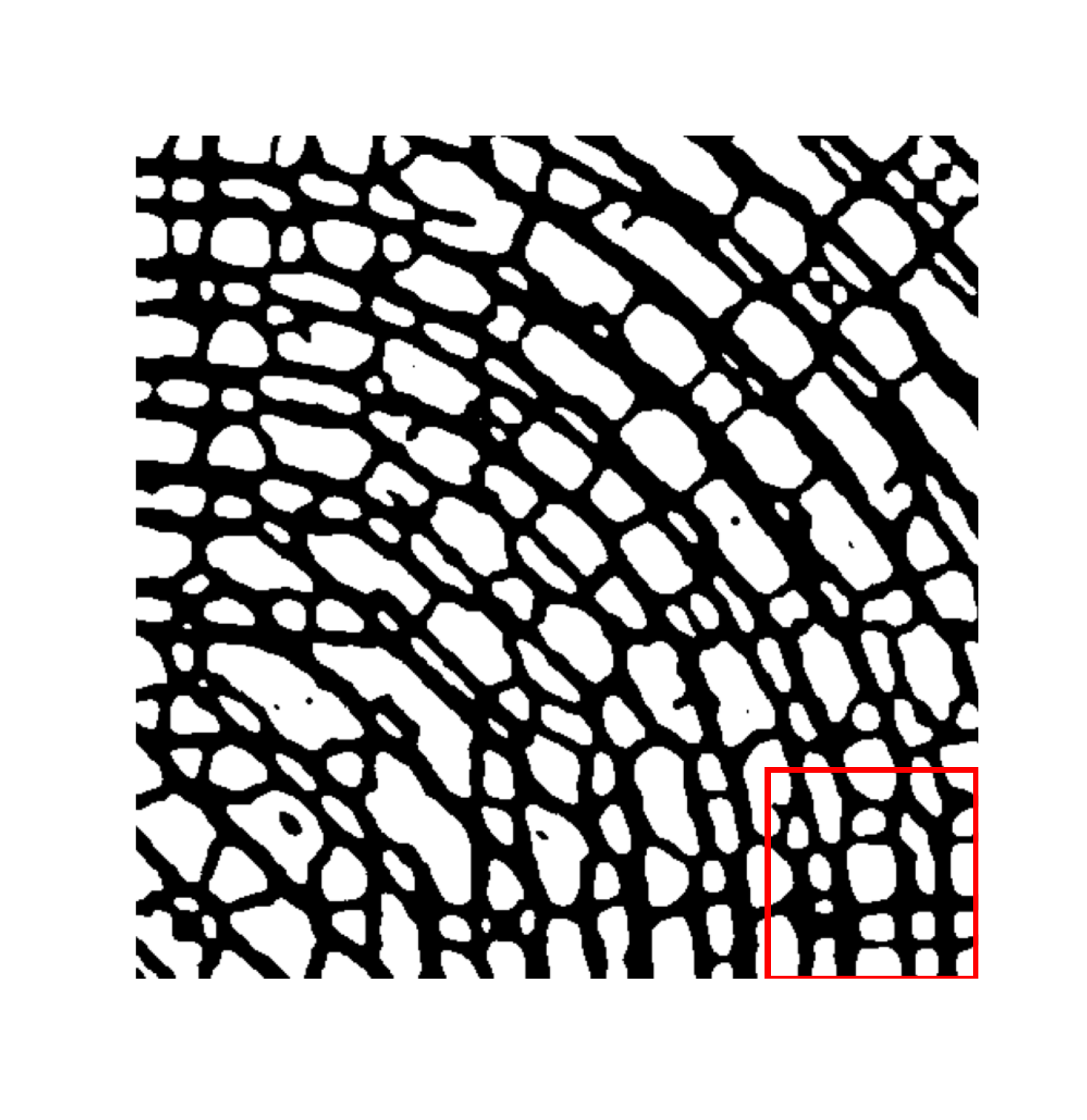}
\caption{Multi-scale structure}
\end{subfigure}
\qquad
\begin{subfigure}[t]{0.3\textwidth}
\centering
\includegraphics[width=\textwidth]{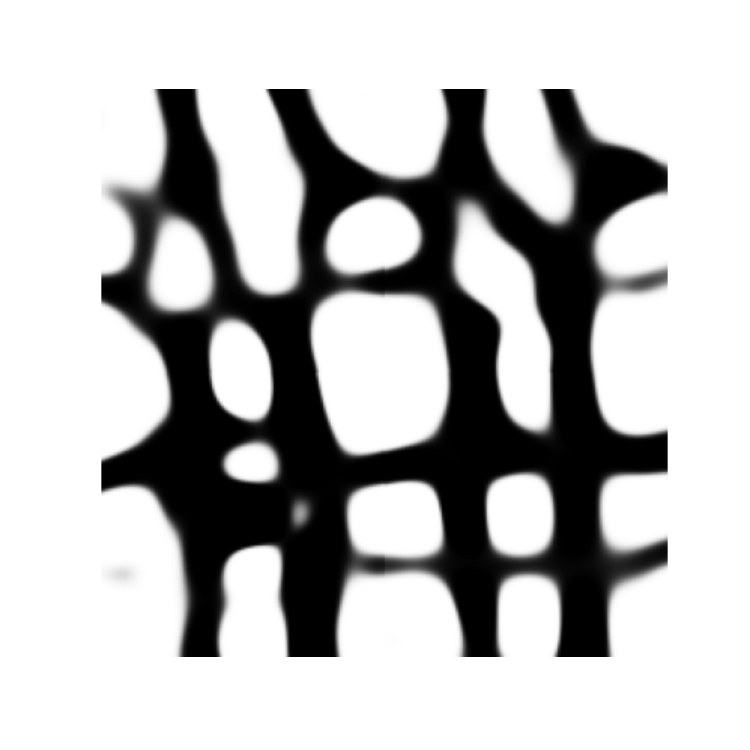}
\caption{With boundary loss}
\end{subfigure}
\qquad
\begin{subfigure}[t]{0.3\textwidth}
\centering
\includegraphics[width=\textwidth]{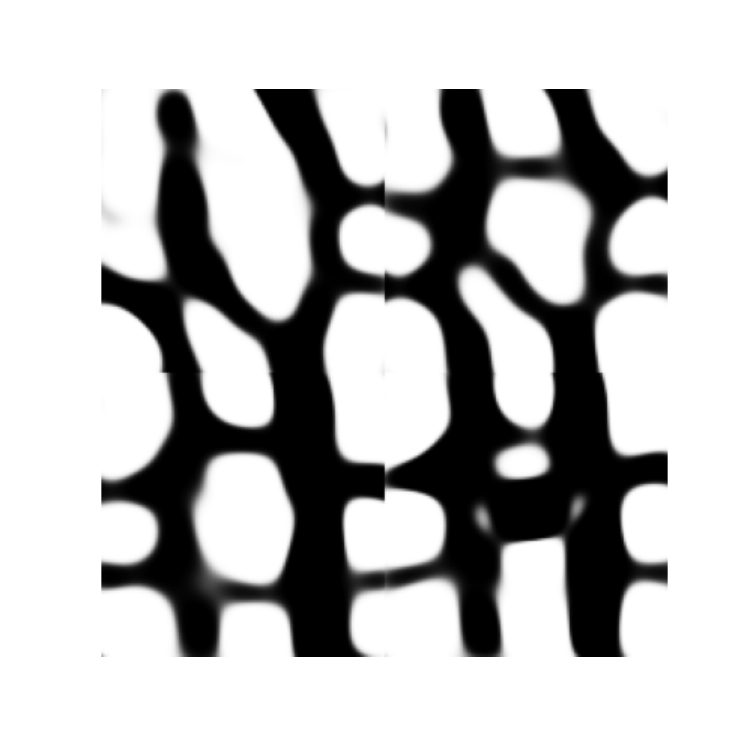}
\caption{Without boundary loss}
\end{subfigure}

\caption{The boundary loss is defined as the L1 loss between the red solid line region and the blue dashed line region in (a). When we zoom into the right bottom corner of the multi-scale structure shown in (b), we can observe that with boundary loss (c), the connection between cells is more continuous when upsampled compared to without boundary loss (d). }

\label{fig:bcloss}
\end{figure*}

\section{PROPOSED METHOD}
In our proposed method, the density distribution of the multi-scale structure is directly represented by the topology neural network. 

\subsection{Neural network-based topology representation}

The topology network $T(\textbf{X})$  (Figure \ref{fig:flowchart}), learns a density field differently as compared to typical topology optimization which represents the density field as a finite element mesh. The topology neural network takes in local domain coordinates $u,w$, as well as the global coordinate $x,y$ for each cell. The global coordinates get concatenated with the local coordinates to form the input to the topology network, $\textbf{X} = [x,y,u,w]$. The local coordinates are normalized between $-0.5$ to $0.5$. The global coordinates are normalized between $-0.5$ to $0.5$ along the longest axis. Given the local and global coordinates, the topology network outputs the density value $\rho$ at each coordinate point. The domain coordinates represent the center of each element in the design domain. During topology optimization, a batch of domain coordinates that correspond to the mesh grid and the corresponding strain energy field is fed into the topology network. The output is then sent to the FE solver. The solver outputs the stiffness objective which is combined with the volume fraction violation as a loss. The loss is then backpropagated to learn the weights of the topology network. 

We implement the neural network in TensorFlow. For the topology network design, we employed a simple architecture that resembles the function expression of $f(x) = \textbf{w}sin(\textbf{k}x+\textbf{b})$. Similar neural network architectures have been used to control the length scale of geometry in topology optimization\cite{Chandrasekhar2021Fourier}. The conditioned domain coordinates are multiplied with a kernel $\textbf{K}$. The kernel $\textbf{K}$ regulates the frequency of the sine function. We add a constant value of 1 to break the sine function's rotation symmetry around the origin. We use a Sigmoid function to guarantee the output is between 0 and 1. The topology network can be formulated as follows:
\begin{equation}
T(\textbf{X}) = \sigma(\textbf{W}\sin(\textbf{K}\textbf{X} +  1))
\end{equation}

\noindent where:

$\textbf{X}$: Domain coordinate input, $\textbf{X}=(x,y,u,w)$ 

$\sigma$: Sigmoid activation function

$\textbf{K}$: Trainable frequency kernels

$\textbf{W}$: Trainable weights

\noindent 

\x{For a typical multi-scale problem, we employ around 5k frequency kernels and a same number of corresponding trainable weights. This means a total of 10k trainable parameters to represent a multi-scale structure.  }

\x{

\subsection{Inverse Homogenization}
We rely on an energy-based homogenization approach to compute the elasticity tensor of each cell and design the material distribution of each cell. A detailed review and accompanying code can be found in the work from Xia and Breitkopf \cite{xia2015design}. Nevertheless, for the sake of clarity, we include a brief review of the formulation of homogenization and inverse homogenization in this section.  

Homogenization aims to evaluate the equivalent constitutive behavior of repetitive microstructures. A compact and easy-to-implement formulation for homogenization is the energy-based approach based on the average stress and strain theorems. The homogenized stiffness tensor $E_{ijkl}^H$ can be approximated in terms of the discretized N mutual energy of elements of the design domain $Y$. 

\begin{equation}
E_{ijkl}^H = \frac{1}{|Y|}\sum_{e=1}^N (\textbf{u}_e^{A(ij)})^T\textbf{k}_e \textbf{u}_e^{A(kl)}
\end{equation}

Once the constitutive stiffness tensor is computed, inverse homogenization refers to the minimization of a combination of the homogenized stiffness tensor. 

\begin{equation}
\begin{aligned}
    \min_{\rho} & \quad c\left(E_{ijkl}^H(\rho)\right) \\
    \text{s.t.} & \quad \mathbf{KU}^{A(kl)} = \mathbf{F}^{(kl)}, \quad k, l = 1, \ldots, d, \\
                & \quad \sum_{e=1}^N v_e \rho_e / |Y| \leq \vartheta, \\
                & \quad 0 \leq \rho_e \leq 1, \quad e = 1, \ldots, N.
\end{aligned}
\end{equation}

The optimization equation involves $ \mathbf{K} $, the global stiffness matrix, while $ \mathbf{U}^{A(kl)} $ and $ \mathbf{F}^{(kl)} $ represent the global displacement vector and the external force vector for the test case $(kl)$, respectively. Here, $ d $ refers to the spatial dimensionality, $ v_e $ is the volume of an individual element, and $ \vartheta $ defines the upper limit for the volume fraction. The objective function $ c(E_{ijkl}^H) $ depends on the homogenized stiffness tensors. For example, the maximization of the bulk modulus can be written with an objective function: 

\begin{equation}
c = -(E_{11} + E_{12} + E_{21} + E_{22})
\label{eq:bulk_modulus}
\end{equation}

}
\subsection{Multi-scale topology optimization with neural networks}

The goal of using the neural network to represent the topology is to enforce cell compatibility. The first mechanism to encourage compatibility is extending the cell boundaries during inverse homogenization, illustrated in Figure \ref{fig:febc_extension}. In 2D, we extend the boundary of each cell by 1.2 times to cover all its 8 neighboring cells. If rotation of the cell is required, the boundary is further extended to 1.6 times to cover all 8 neighboring cells. For the extended regions that are in neighboring cells, the domain coordinates use the local and global coordinates for neighboring cells. We use 4 node square elements for finite element analysis. Within the boundary, we sample the domain at uniform intervals that directly correspond with the finite element mesh. During training, iterative updates to the design domain are carried out on these sampling points. 

The second mechanism to encourage cell compatibility is boundary loss. Once we upsample the input coordinate to obtain geometry at finer resolution, we observe that there are still small inconsistent transitions between cell boundaries which can be seen in Figure \ref{fig:bcloss}. Therefore, a boundary loss is added to improve the consistency between cells. 

The boundary loss is formulated as the L1 loss between the outer edge of the neighboring cells and the outer edge of the center cell. 

\begin{equation}
\mathcal{L}_{bc} = \frac{1}{n}\sum^n |T(X_{neigh}) - T(X_{center})|
\end{equation}

As illustrated in Figure \ref{fig:bcloss}, the outer region $T(X_{neigh})$ for the 8 neighboring cells consists of the area highlighted in solid red lines and the outer region for the center cell $T(X_{center})$ is highlighted in blue dashed lines. The extent of the sampling for each cell is also expanded by 1.2 times. For the regular sampling for the continuous multi-scale structure, each local coordinate is normalized between -0.5 and 0.5. During the computation of the boundary loss, each cell is further expanded by 1.2 times such that the local coordinates are in the range of -0.6 to 0.6. \x{We choose the extension ratio based on heuristics such that a small portion of the neighboring cells is covered while not being too big that most of the FE simulation patch consists of neighboring cells instead of the center cell. } During the optimization, the boundary loss is turned on after 50 epochs.

During optimization, the topology network outputs the density value at the center for each element. These density values are then sent to the finite element solver to calculate the homogenized stiffness tensor based on the SIMP interpolation. We adopted the educational MATLAB code by Xia and Breitkopf \cite{xia2015design} for inverse homogenization and programmed the FE code in Python to connect with the neural network that is implemented in TensorFlow.

The homogenization is based on an energy-based approach where unit test strains are directly imposed on the boundaries of the base cell. In order to achieve a different ratio of the stiffness tensor target, a weight $w$ is assigned to each of the components of the stiffness tensor. 

\begin{equation}
c = - (w_{11}E_{11} + w_{12}E_{12} + w_{13}E_{13} + w_{22}E_{22} + w_{23}E_{23} + w_{33}E_{33})
\end{equation}

The finite element solver is treated as a black box within the neural network. It takes in the density of each element and outputs the homogenized stiffness tensor and the sensitivity for each element with respect to the stiffness tensor. Variables that are being optimized are the weights $\textbf{W}$ and kernels $\textbf{K}$ of the neural network. Adam\cite{kingma2014adam} is used to train the neural network \x{with a learning rate of 0.002}. The constrained optimization problem needs to be transferred into an unconstrained minimization problem for neural networks. The combined loss function is 

\begin{equation}
\mathcal{L} = \frac{1}{n}\sum^n \frac{c_i}{c_{0,i}} + \alpha(\frac{V_i}{V^*_i}-1)^2 + \beta\mathcal{L}_{bc,i}
\end{equation}

\begin{figure*}

\centering
\includegraphics[width=\textwidth]{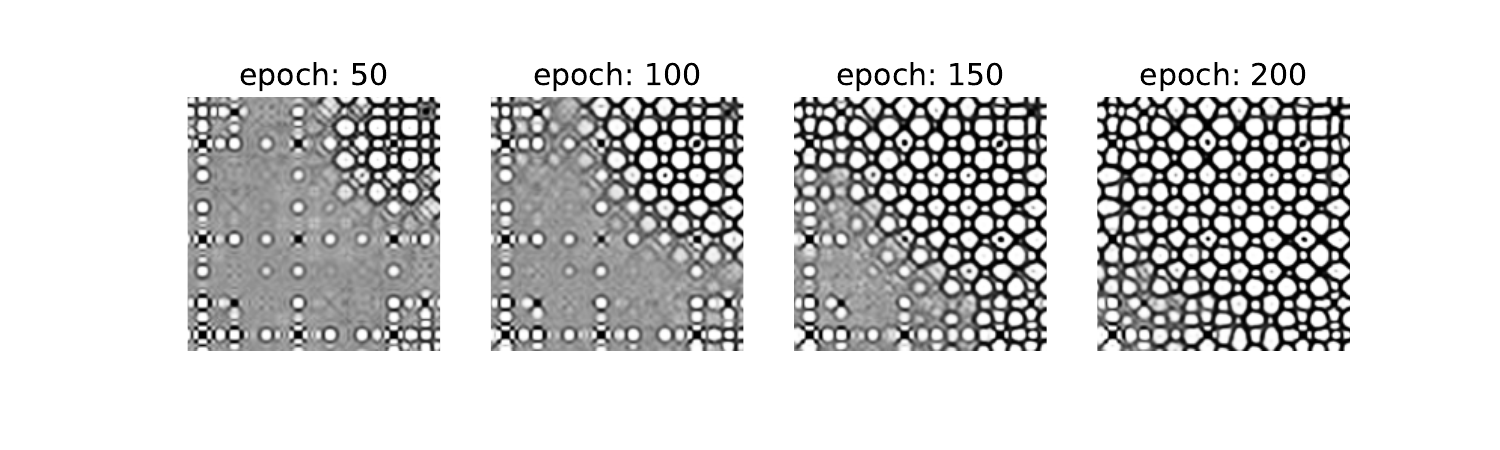}

\caption{The convergence history of the multi-scale structure from the neural network. Microstructures emerge as clusters and from these clusters gradually propagate through the design domain.}

\label{fig:ct_convergence}
\end{figure*}

The objective of stiffness tensor $c_i$ is calculated for each cell $i$ and normalized against the \x{stiffness objective} $c_{0,i}$ which is calculated from a homogenized density field with volume fraction of $V^*_i$. $\alpha$ is the penalty factor volume fraction $V_i$ of each cell $i$. Starting from the first epoch, it gradually increases to 100 at the end of the optimization. $\beta$ is the penalty factor for the boundary loss $\mathcal{L}_{bc}$ which remains 0 until epoch 50 and gradually increases to 1 at the end of the optimization. We choose epoch 50 as we can observe some microstructure starts to appear while the domain is still largely not converged. One example of convergence history is plotted in Figure \ref{fig:ct_convergence}. A similar loss function for a single scale has also been used by Chandrasekhar and Suresh \cite{Chandrasekhar2021} for compliance minimization and volume fraction constraint with neural networks.

\begin{figure*}

\centering

\begin{subfigure}[t]{0.3\textwidth}
\centering
\includegraphics[width=0.85\textwidth]{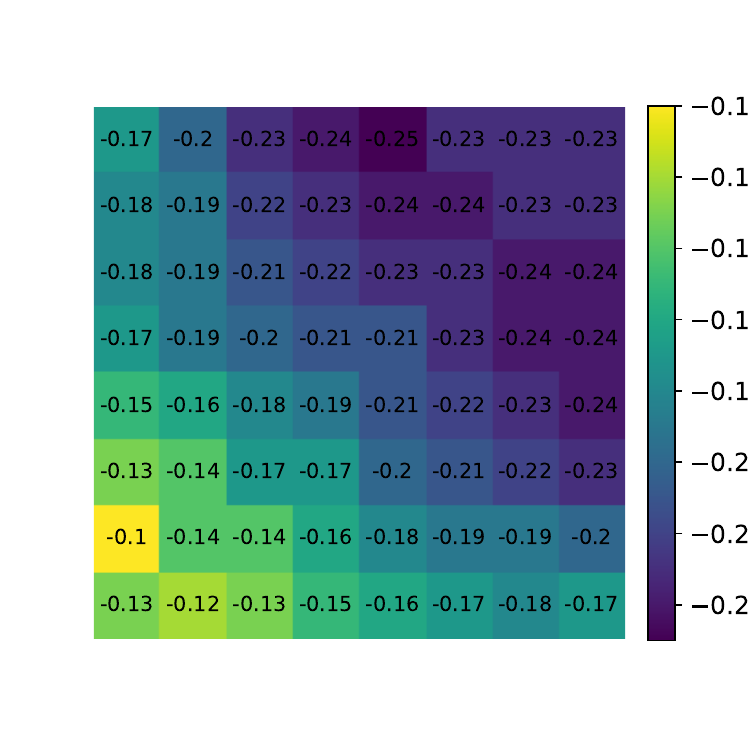}
\caption{Objective function value}
\end{subfigure}
\qquad
\begin{subfigure}[t]{0.3\textwidth}
\centering
\includegraphics[width=\textwidth]{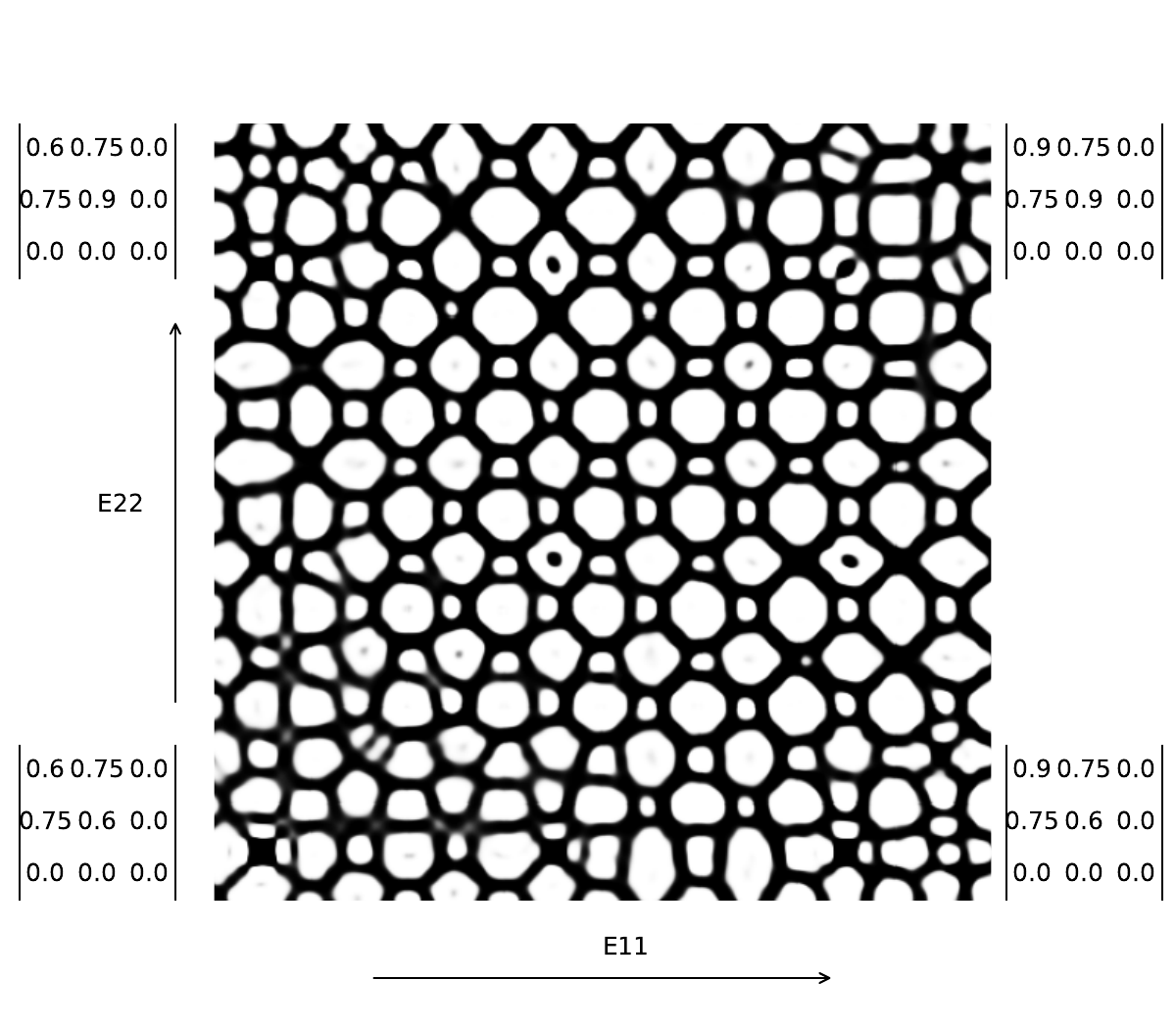}
\caption{Multi-scale structure}
\end{subfigure}

\caption{Given a changing stiffness tensor weights, we can observe in (a) that, as the combined weights increase to the top right corner, the objective function value for each cell is also smaller. From the converged multi-scale structure, we can observe good connectivity between the cells. }

\label{fig:ct_field}
\end{figure*}
\begin{figure*}

\centering

\begin{subfigure}[t]{0.3\textwidth}
\centering
\includegraphics[width=0.85\textwidth]{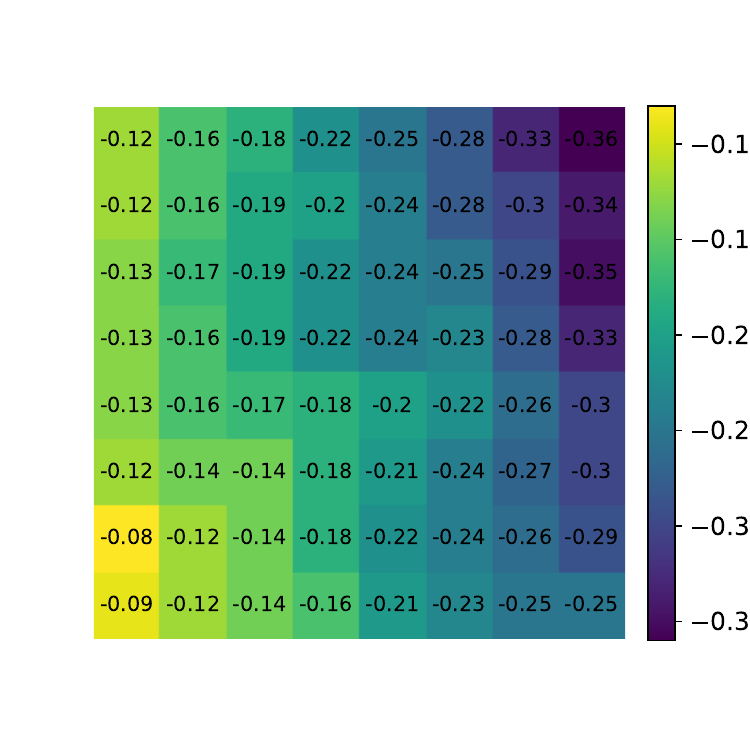}
\caption{Objective function value}
\end{subfigure}
\qquad
\begin{subfigure}[t]{0.3\textwidth}
\centering
\includegraphics[width=\textwidth]{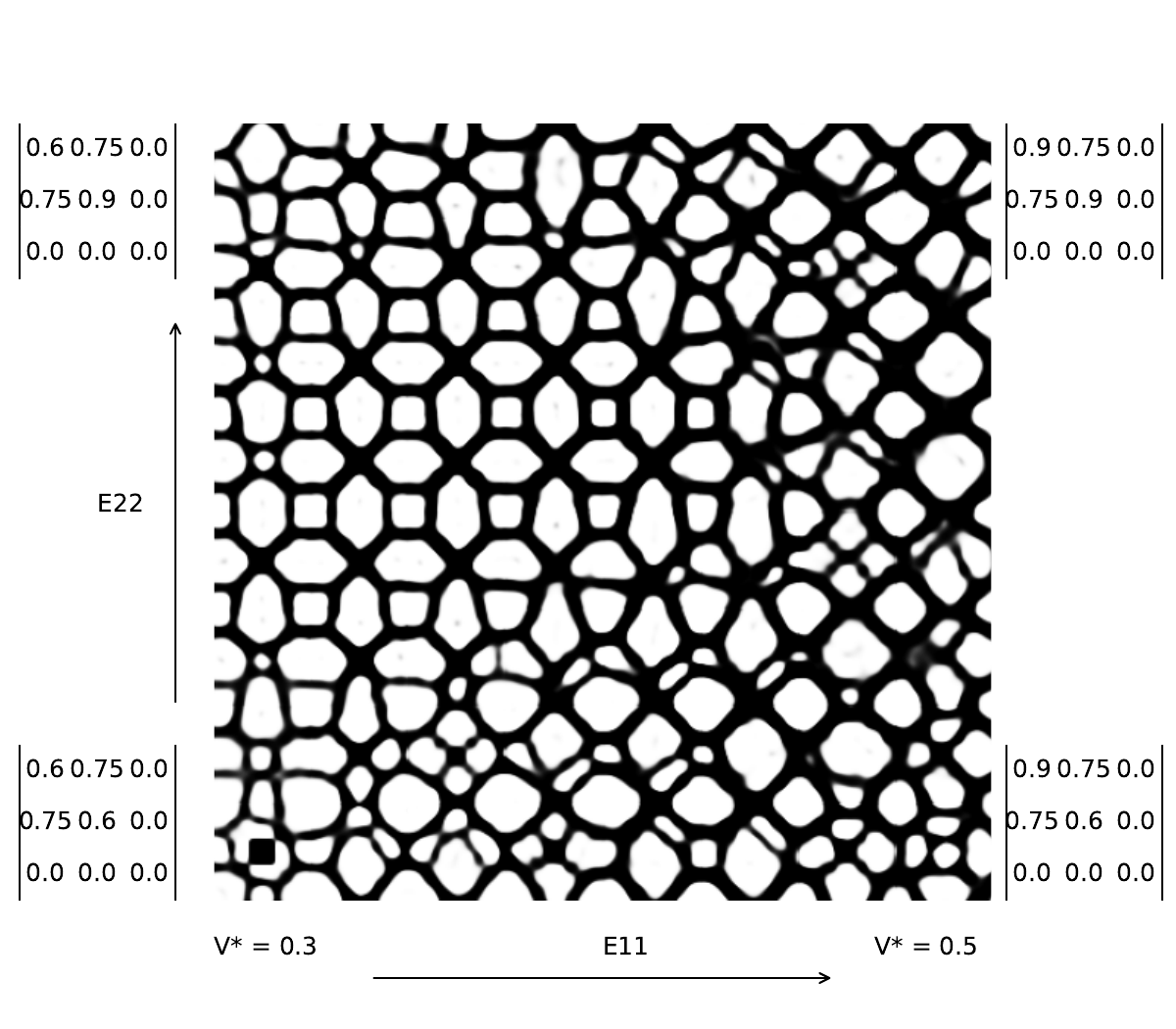}
\caption{Multi-scale structure}
\end{subfigure}

\caption{The volume fraction target is varied for each cell along the horizontal direction. Good compatibility is observed as the microstructure edges gradually thicken. The objective value is also lower with higher volume fraction as the stiffness tensor is greater for cells with higher volume fraction. }

\label{fig:ctvf_field}
\end{figure*}
\begin{figure*}

\centering

\begin{subfigure}[t]{0.3\textwidth}
\centering
\includegraphics[width=0.85\textwidth]{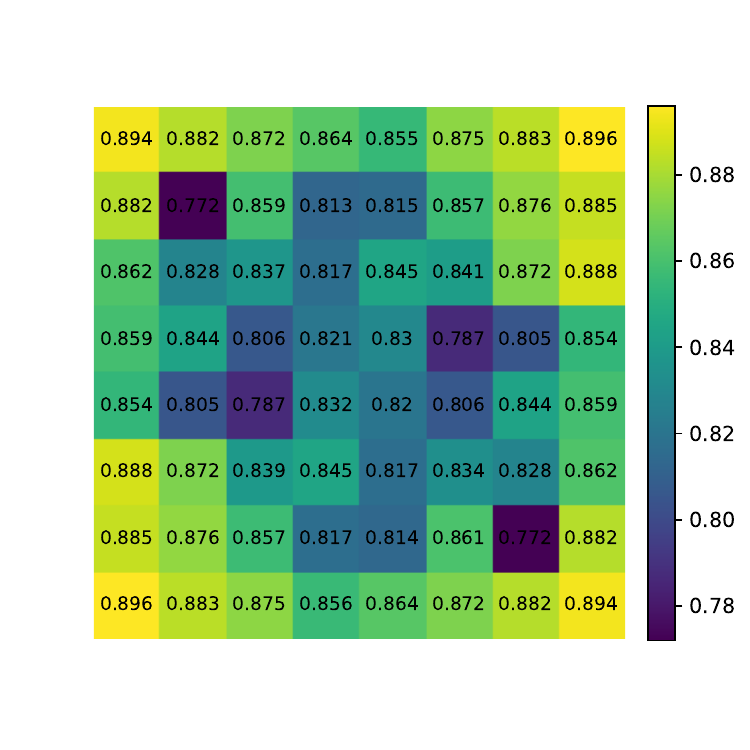}
\caption{Bulk modulus versus HS upper bound ratio.  }
\end{subfigure}
\qquad
\begin{subfigure}[t]{0.3\textwidth}
\centering
\includegraphics[width=0.85\textwidth]{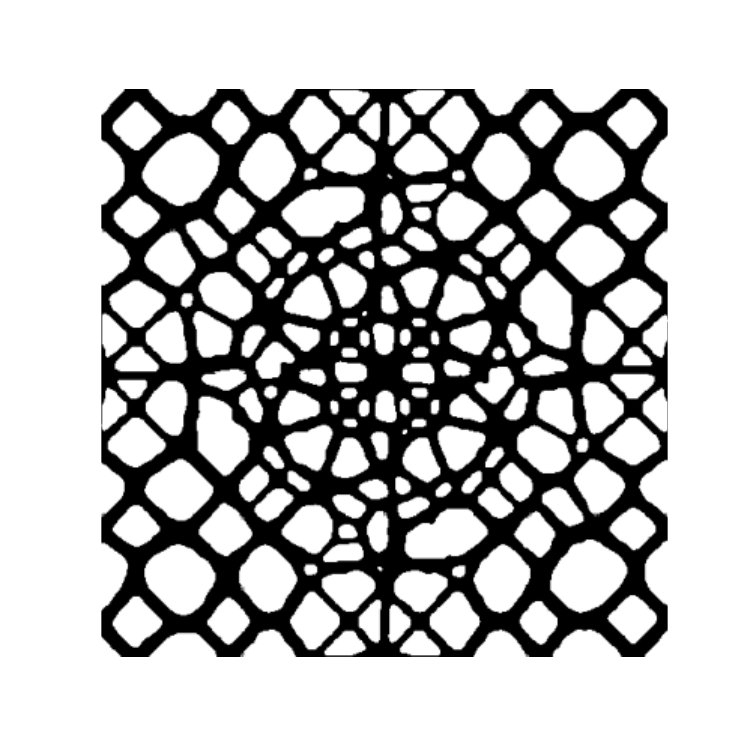}
\caption{Multi-scale structure}
\end{subfigure}

\begin{subfigure}[t]{0.3\textwidth}
\centering
\includegraphics[width=0.85\textwidth]{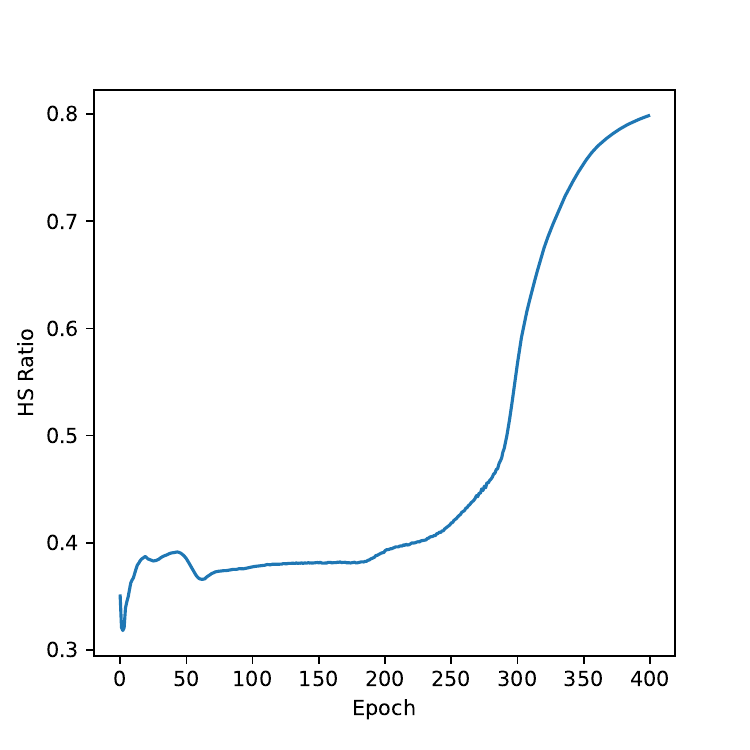}
\caption{Convergence history}
\end{subfigure}
\qquad
\begin{subfigure}[t]{0.3\textwidth}
\centering
\includegraphics[width=0.85\textwidth]{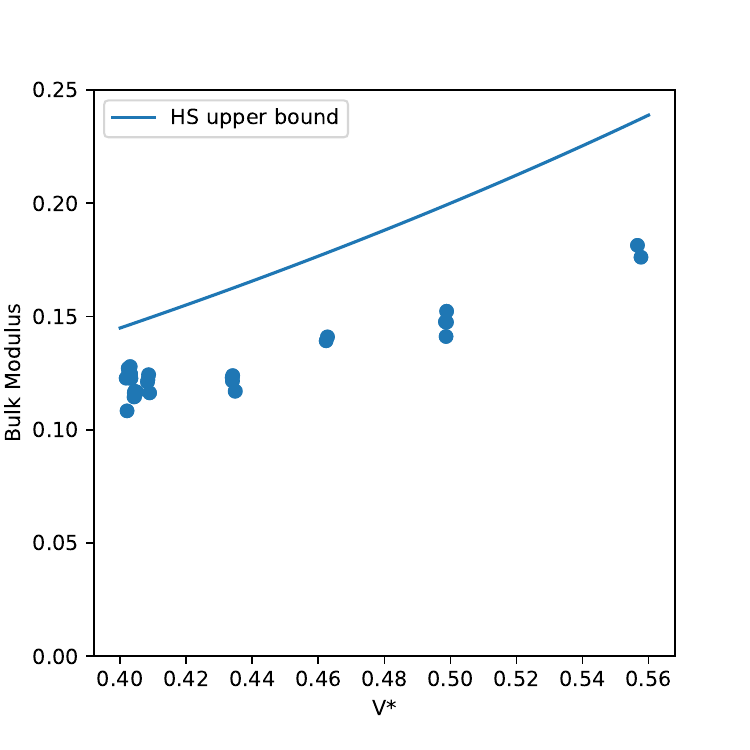}
\caption{Bulk modulus vs volume fraction}
\end{subfigure}

\caption{We varied the volume fraction between 0.4 to 0.56 and maximized the bulk modulus. Comparing the bulk modulus from the HS upper bound, shown in (a) for each cell, we achieved an average of 84.9$\%$. The convergence history shown in (c) and the bulk modulus vs volume fraction plot in (d) were using the SIMP interpolation of Young's modules to compare against the HS upper bound. The comparison in (a) uses a binarized density field with volume fraction recalculated for each cell after binarization.  }

\label{fig:bulk_field}
\end{figure*}
\begin{figure*}

\centering

\begin{subfigure}[t]{0.45\textwidth}
\centering
\includegraphics[width=0.6\textwidth]{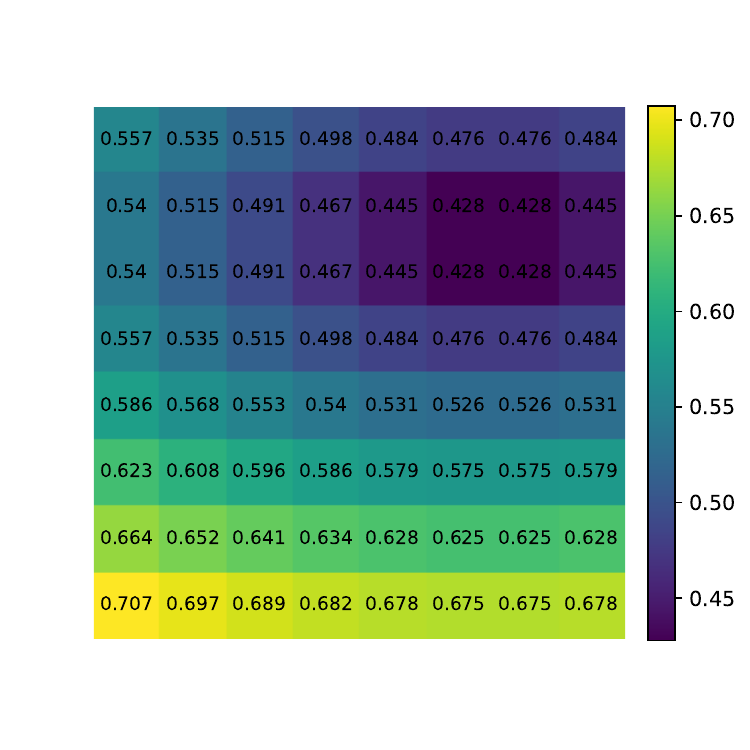}
\caption{Volume fraction target}
\end{subfigure}
\qquad
\begin{subfigure}[t]{0.45\textwidth}
\centering
\includegraphics[width=0.6\textwidth]{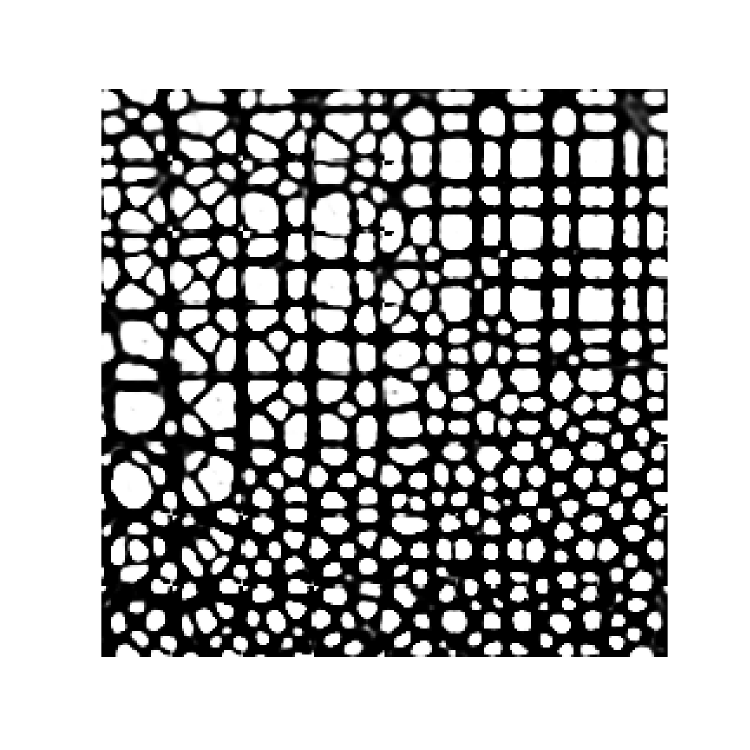}
\caption{Full batch, 449s, 85$\%$}
\end{subfigure}

\begin{subfigure}[t]{0.45\textwidth}
\centering
\includegraphics[width=0.6\textwidth]{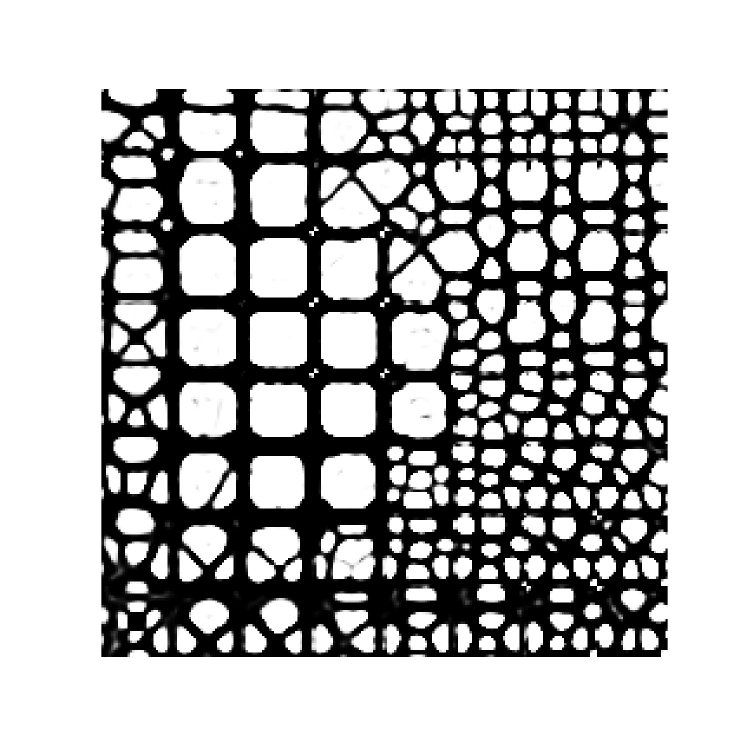}
\caption{1/2 mini-batch, 465s, 86$\%$}
\end{subfigure}
\qquad
\begin{subfigure}[t]{0.45\textwidth}
\centering
\includegraphics[width=0.6\textwidth]{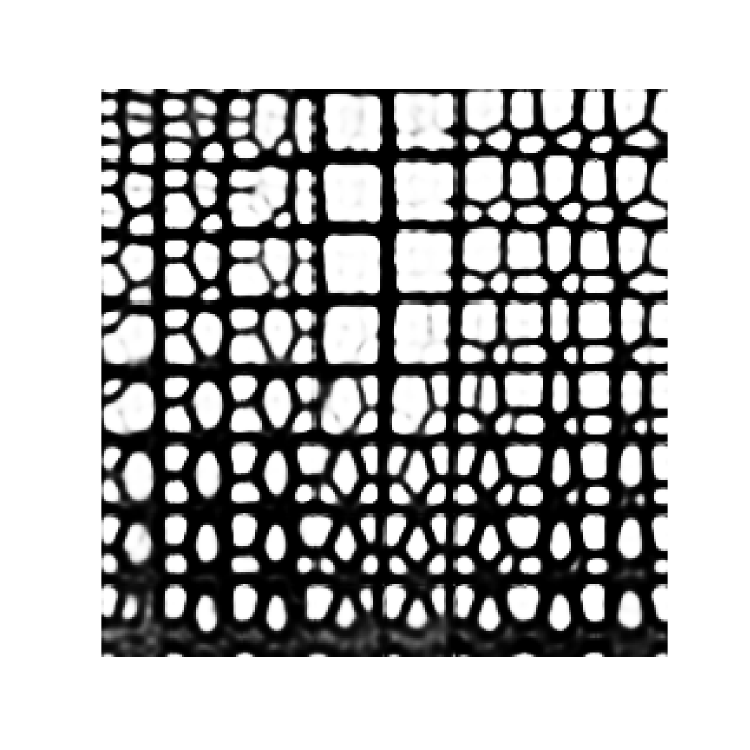}
\caption{1/4 mini-batch, 500s, 85$\%$}
\end{subfigure}

\begin{subfigure}[t]{0.45\textwidth}
\centering
\includegraphics[width=0.6\textwidth]{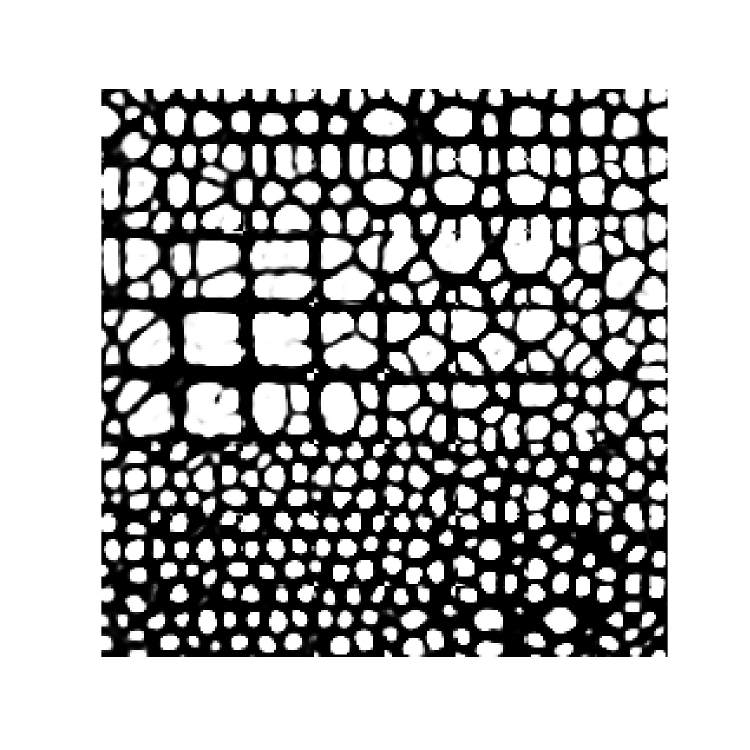}
\caption{1/2 mini-epoch, 223s, 83$\%$}
\end{subfigure}
\qquad
\begin{subfigure}[t]{0.45\textwidth}
\centering
\includegraphics[width=0.6\textwidth]{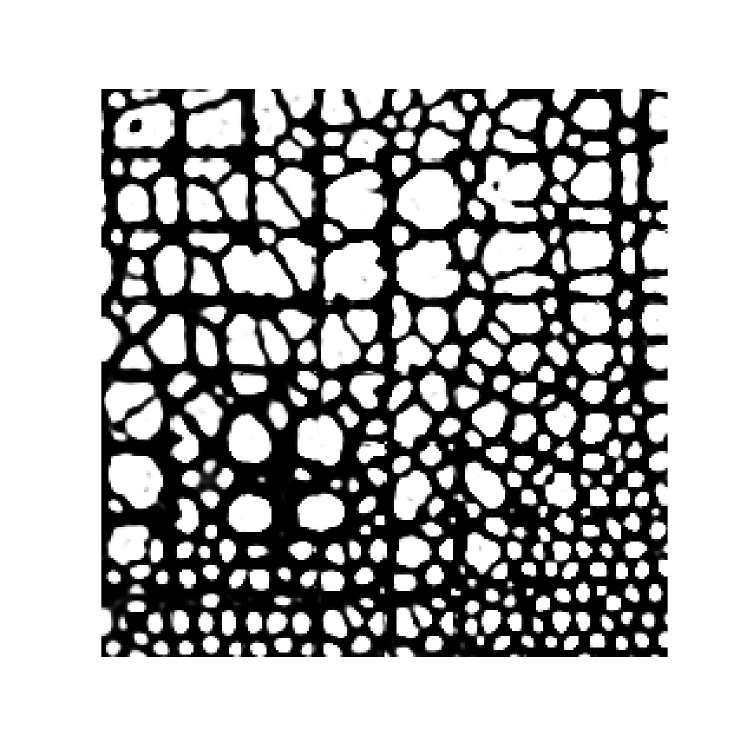}
\caption{1/4 mini-epoch, 123s, 80$\%$}
\end{subfigure}

\caption{Comparing the runtime and the percentage of Hashin–Shrickman upper bound reached given different batch sizes and total epochs. For runs in (c,d), we configured a mini-batch where each batch is 1/2 or 1/4 of the total number of cells. We observed stable convergence with slightly increased runtime due to the increased backpropagation count. For runs in (e,f), we explored if the interpretation capability of the neural network allows us to use a subset of cells every epoch. We use 1/2 or 1/4 of the cells per epoch, which we call mini-epoch. The runtime showed almost half and a quarter of the runtime compared to the full batch, with a small penalty in mechanical performance.   }

\label{fig:batch_hs}
\end{figure*}

\section{RESULTS AND DISCUSSIONS}
We set up problems with varying stiffness tensor targets, changing volume fractions, and rotation of stiffness tensor targets. These problems aim to test the ability of the neural network-based multi-scale optimization to design compatible microstructure. All experiments are run on a PC with i7-12700K as the processor, 32 GB of RAM, and Nvidia RTX3080 GPU. 

\subsection{Multi-scale topology optimization with varying stiffness tensor}
We first evaluate the effectiveness of our approach with changing stiffness tensor targets. The stiffness tensor target is formulated as a combination of varying weights on the compliance calculated for each cell. For example, the bottom left side of Figure \ref{fig:ct_field} targets a stiffness tensor with the ratio of: $\begin{bmatrix} 0.6 & 0.75 & 0.0 \\ 0.75 & 0.6 & 0.0 \\ 0.0 & 0.0 & 0.0 \end{bmatrix}$ is written in as $c = - (0.6E_{11} + 1.5E_{12} + 0.6E_{22})$. 

The example in Figure \ref{fig:ct_field} consists of 8$\times$8 cells on the macro-scale, 30$\times$30 elements on the micro-scale, and a uniform volume fraction target of 0.4. The stiffness tensor weights are configured such that in the horizontal direction $E_{11}$ is increased. In the vertical direction $E_{22}$ is increased. We run the optimization for 300 epochs and plot out a snapshot of the multi-scale structure during training in Figure \ref{fig:ct_convergence}. The optimization takes 9 minutes to run. During optimization, we observe that cells converge in clusters. On the top right corner of Figure \ref{fig:ct_convergence} is a large cluster where the cells converged faster. As the optimization progresses, the front with converged microstructure gradually progresses to the bottom left. This could indicate that the neural network is using cells that are closer in global coordinates to interpolate. 

In addition to changing the stiffness tensor weights, the volume fraction can also be varied. Compared to the example in Figure \ref{fig:ct_field}, the example in Figure \ref{fig:ctvf_field} has the volume fraction target increasing from 0.3 to 0.5 in the horizontal axis. We can observe the thickness of the multi-scale structure smoothly increased to accommodate for the changing volume fraction. The compliance evaluated in each cell is plotted in Figure \ref{fig:ctvf_field} (a), where it is lower with a higher volume fraction. 

We then configure a problem with varying volume fraction and bulk modulus maximization. This allows us to compare the performance of the theoretical Hashin–Shrickman (HS) upper bound \cite{hashin1963variational}. To maximize for the bulk modulus, the weightage on the stiffness tensor is configured as in equation \ref{eq:bulk_modulus}. The converged multi-scale structure is plotted in Figure \ref{fig:bulk_field} (b). The volume fraction target in the center is 0.6 and is gradually decreased to 0.4 on the outer edges. We threshold the density field at 0.4. The bulk modulus of each cell and the corresponding HS upper bound are calculated with the thresholded density field. The ratio between the computed bulk modulus versus the HS upper bound is then plotted in \ref{fig:bulk_field} (a). On average, across 64 total cells and between 0.4 and 0.56 volume fraction, we achieved 84.9$\%$ of the HS upper bound for bulk modulus. This is slightly lower compared to the work by Sridhara et al.\cite{sridhara2022generalized} where for single microstructure design by the neural network, they achieved 90$\%$ of the HS upper bound in the volume fraction range. Slightly higher restrictions on design freedom can be expected with the connectivity constraints posed. We note that no explicit symmetry constraint is applied to this example. The resulting structure demonstrated 2-fold rotation symmetry. While mirror symmetry is missing, this is attributed to the architecture of the neural network. Inputting global coordinates from the I and III or II and IV quadrants leads to the same result. For results where mirror symmetry is desired, explicit mirroring of the global coordinates needs to be performed. Previous works with neural network-based topology optimization also required such mirroring \cite{Chandrasekhar2021}. 

\x{
Two potential issues with the previous experiments are the linearly increasing memory consumption and longer runtime with full batches. Therefore, we explored mini-batches to reduce the number of cells seen at every backpropagation step and mini-epochs for which we only let the optimizer see a subset of cells every epoch to exploit the interpretability of neural networks. We compared runtime and the percentage of the Hashin–Shrickman upper bound achieved under different batch sizes and total epochs. In Figure \ref{fig:batch_hs} experiments (c, d), we used mini-batches where each batch contained 1/2 or 1/4 of the total cells. This setup resulted in stable convergence but slightly increased runtime due to more frequent backpropagation. In experiments (e, f), we tested whether the neural network's interpretability allowed using a subset of cells per epoch, referred to as a mini-epoch. Since neighboring cells share similar global coordinates, we observed that during convergence, the neural network seems to leverage the knowledge of neighboring cells to infer the overall microstructure, effectively accelerating convergence. This behavior suggests that the network learns to generalize the spatial relationships and extrapolate the microstructure patterns from a subset of the data. Additionally, as neural networks are designed to approximate continuous functions, the learned representation appears to smooth out local variations and capture global trends, further enhancing training efficiency with reduced data. With 1/2 or 1/4 of the cells per mini-epoch, the runtime was reduced to nearly half or a quarter of the full-batch runtime, with only a slight trade-off in mechanical performance. 

}

\begin{figure*}

\centering

\begin{subfigure}[t]{0.3\textwidth}
\centering
\includegraphics[width=\textwidth]{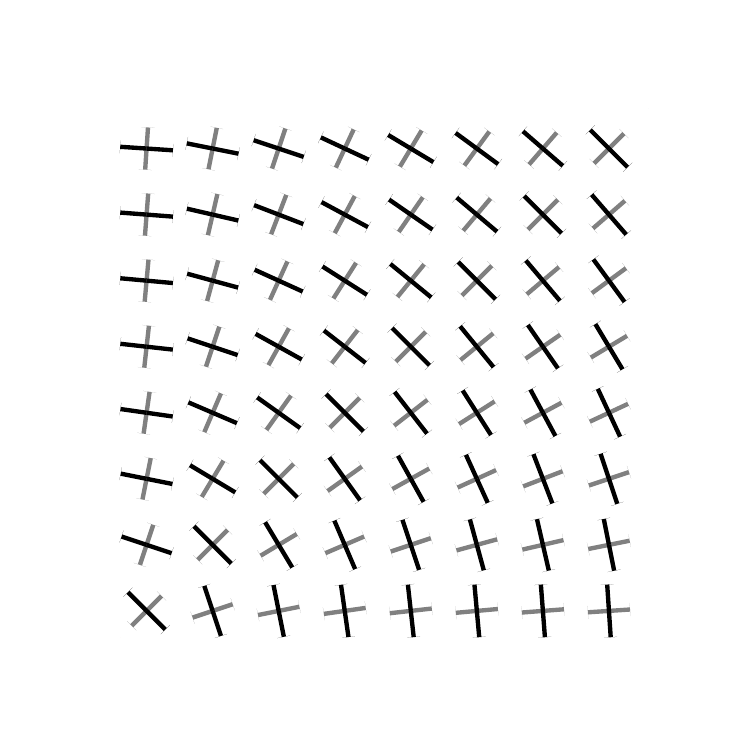}
\caption{Stiffness tensor orientation}
\end{subfigure}
\qquad
\begin{subfigure}[t]{0.3\textwidth}
\centering
\includegraphics[width=\textwidth]{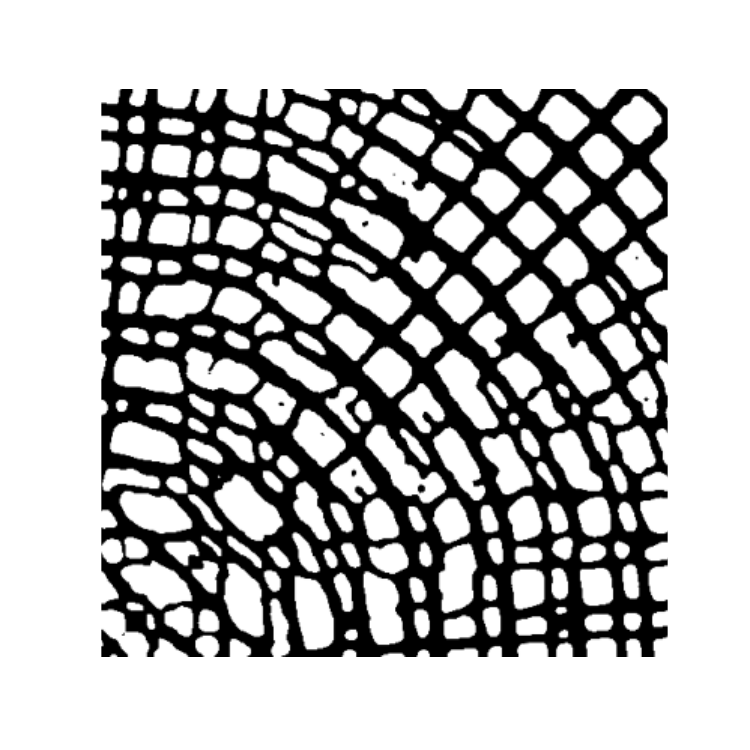}
\caption{Rotating geometry}
\end{subfigure}
\qquad
\begin{subfigure}[t]{0.3\textwidth}
\centering
\includegraphics[width=\textwidth]{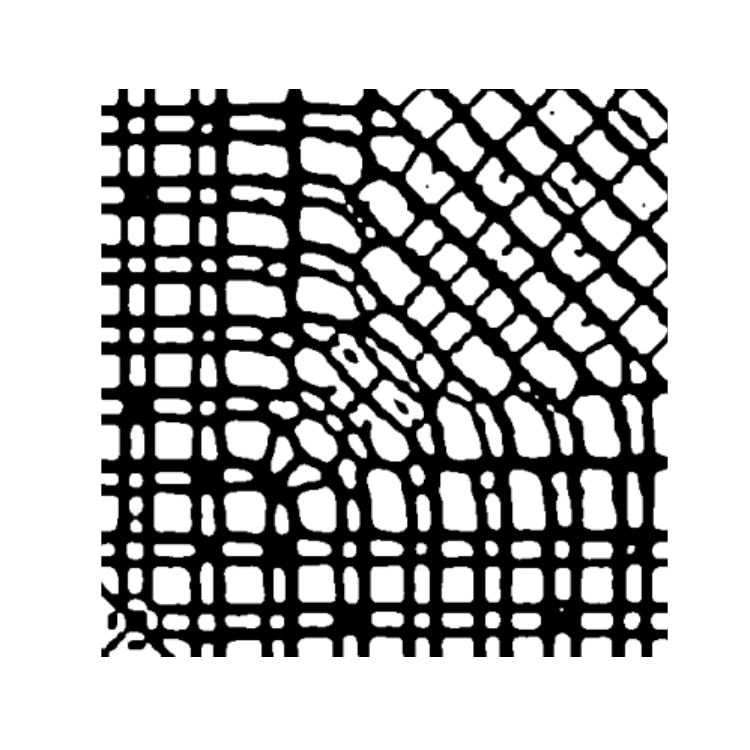}
\caption{Rotating stiffness tensor}
\end{subfigure}

\begin{subfigure}[t]{0.3\textwidth}
\centering
\includegraphics[width=\textwidth]{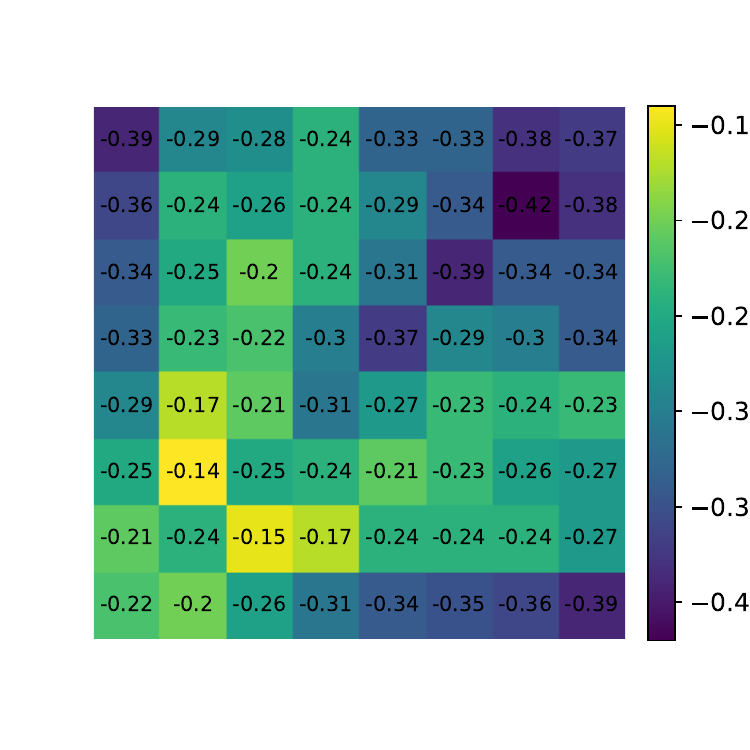}
\caption{Average objective: -0.28}
\end{subfigure}
\qquad
\begin{subfigure}[t]{0.3\textwidth}
\centering
\includegraphics[width=\textwidth]{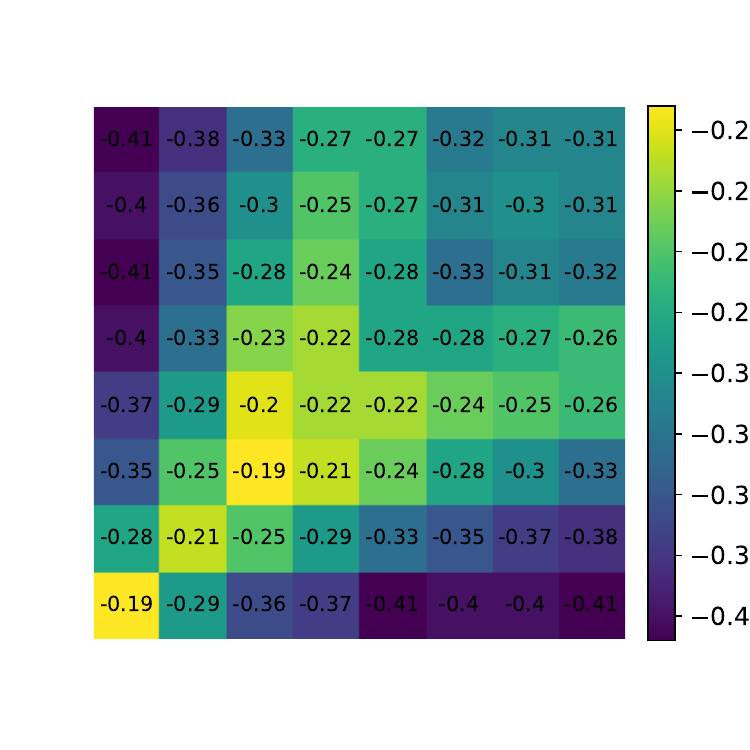}
\caption{Average objective: -0.30}
\end{subfigure}

\caption{(a) The stiffness tensor field provided as input. (b-c) Rotation of the geometry versus rotation of the stiffness tensor. Rotating the geometry results in a microstructure that more closely follows the prescribed rotation. In contrast, rotating the stiffness tensor causes a more abrupt transition.}

\label{fig:rb_field}
\end{figure*}
\begin{figure*}

\centering

\begin{subfigure}[t]{0.3\textwidth}
\centering
\includegraphics[width=0.85\textwidth]{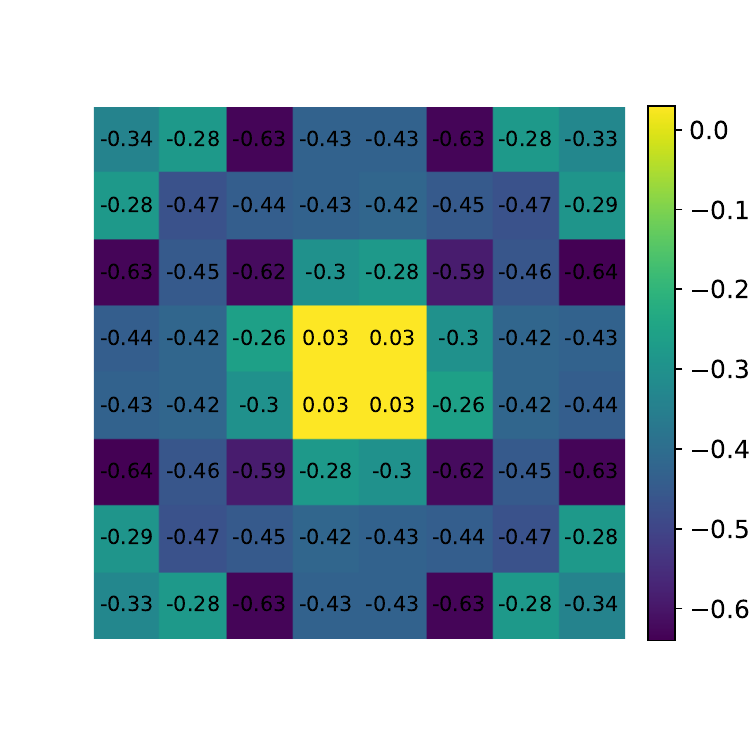}
\caption{Poisson's ratio}
\end{subfigure}
\qquad
\begin{subfigure}[t]{0.3\textwidth}
\centering
\includegraphics[width=0.85\textwidth]{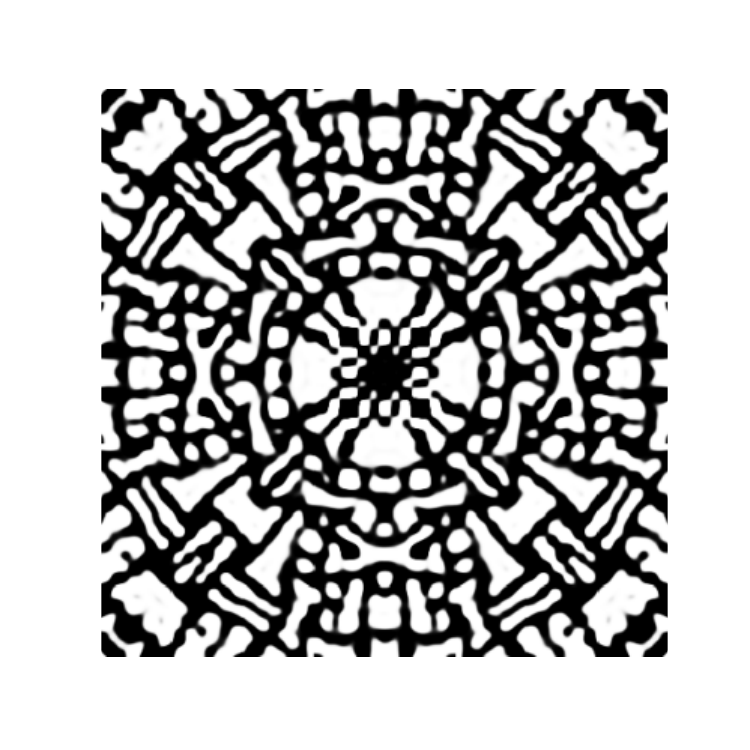}
\caption{Multi-scale structure}
\end{subfigure}

\caption{The figure demonstrates strong connectivity, with each cell rotated around the center. Except for the four central cells, the Poisson's ratio of all other cells is negative.}

\label{fig:rp_field}
\end{figure*}

\begin{figure*}
\centering

\begin{subfigure}[t]{0.25\textwidth}
\centering
\includegraphics[width=\textwidth]{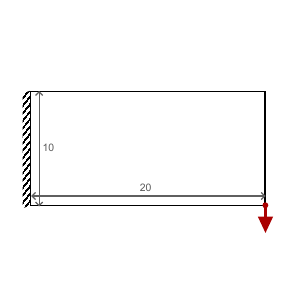}
\caption{}
\end{subfigure}
\qquad
\begin{subfigure}[t]{0.25\textwidth}
\centering
\includegraphics[width=\textwidth]{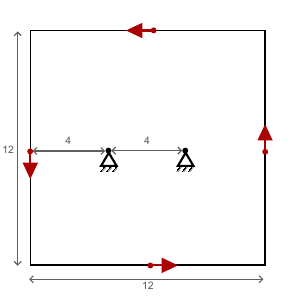}
\caption{}
\end{subfigure}
\qquad
\begin{subfigure}[t]{0.25\textwidth}
\centering
\includegraphics[width=\textwidth]{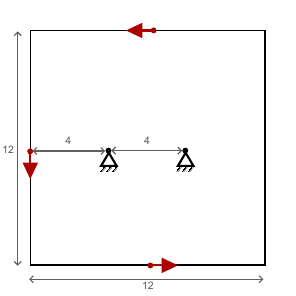}
\caption{}
\end{subfigure}

\begin{subfigure}[t]{0.25\textwidth}
\centering
\raisebox{0.3\height}{ 
    \includegraphics[width=\textwidth]{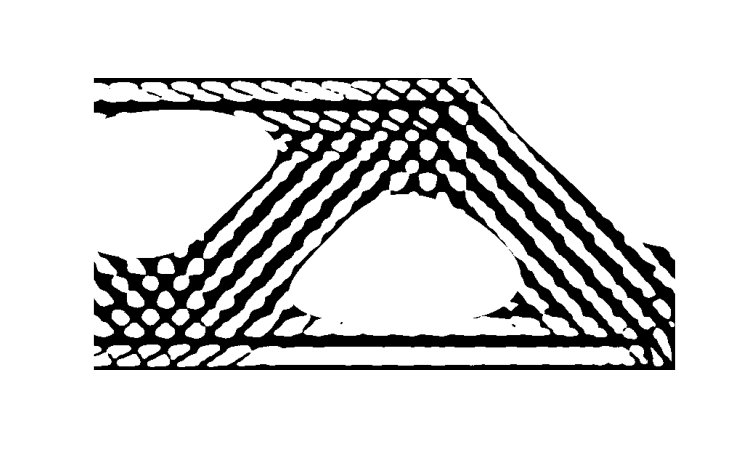}
}
\caption{}
\end{subfigure}
\qquad
\begin{subfigure}[t]{0.25\textwidth}
\centering
\includegraphics[width=\textwidth]{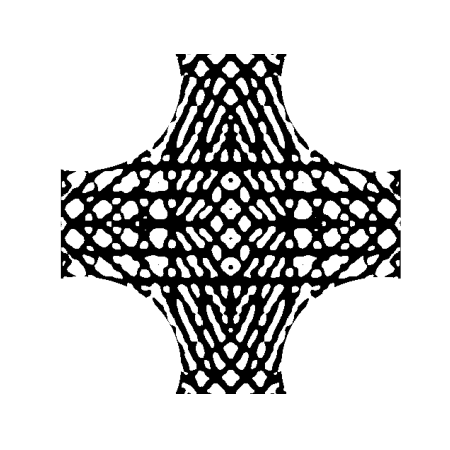}
\caption{}
\end{subfigure}
\qquad
\begin{subfigure}[t]{0.25\textwidth}
\centering
\includegraphics[width=\textwidth]{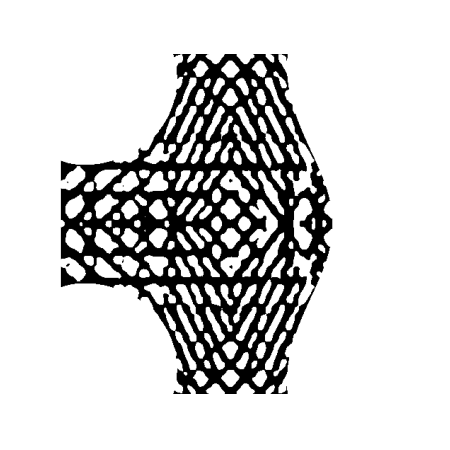}
\caption{}
\end{subfigure}

\caption{(a-c) Boundary conditions for the multiscale topology optimization examples. (d-f) Converged multiscale structures. The neural network-based representation of the macro-scale design field enables upsampling, allowing for a smoother extraction of microstructure boundaries. The resulting multiscale structures demonstrate good cell compatibility. Both macro and micro scale structure reached within 10\% error of their volume fraction target. }
\label{fig:ms}
\end{figure*}

\begin{figure*}

\centering

\begin{subfigure}[t]{0.3\textwidth}
\centering
\includegraphics[width=\textwidth]{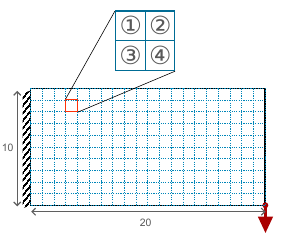}
\caption{}
\end{subfigure}
\qquad
\begin{subfigure}[t]{0.3\textwidth}
\centering
\includegraphics[width=\textwidth]{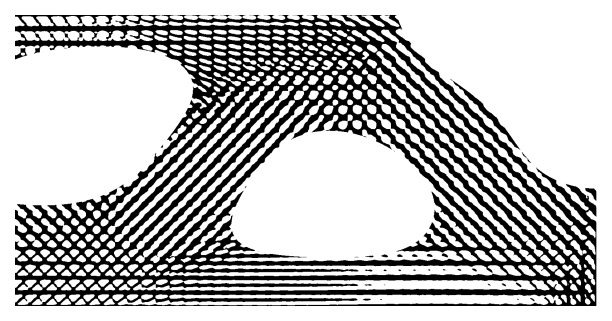}
\caption{}
\end{subfigure}

\begin{subfigure}[t]{0.3\textwidth}
\centering
\includegraphics[width=\textwidth]{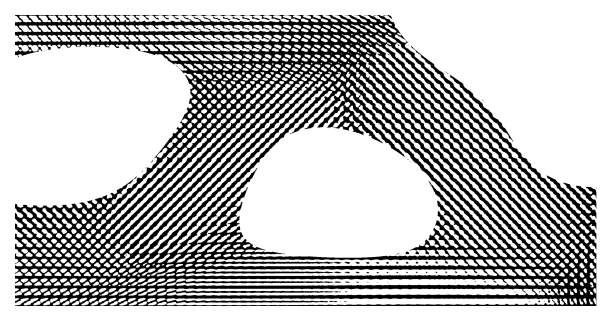}
\caption{}
\end{subfigure}
\qquad
\begin{subfigure}[t]{0.2\textwidth}
\centering
\includegraphics[width=\textwidth]{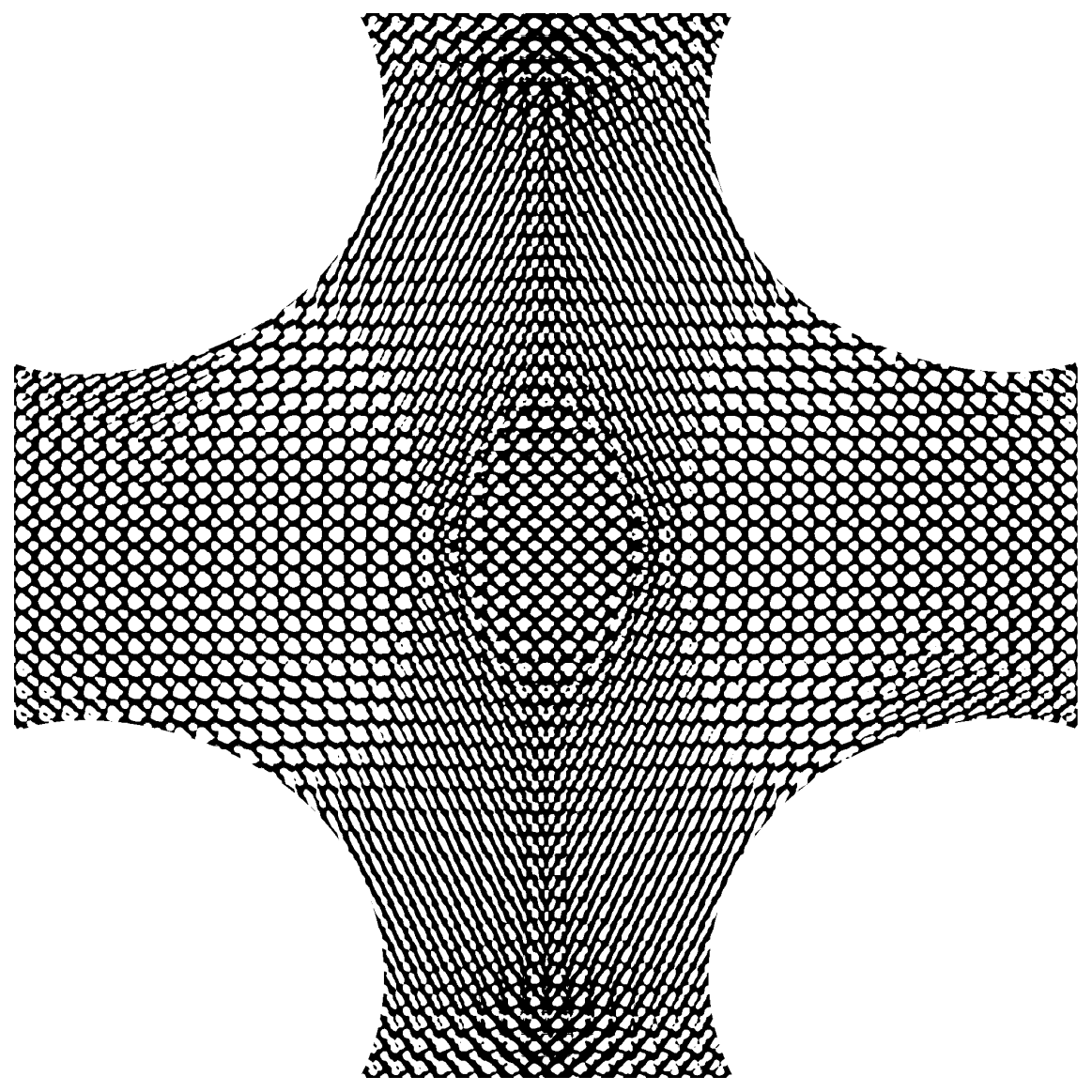}
\caption{}
\end{subfigure}

\begin{subfigure}[t]{0.6\textwidth}
\centering
\includegraphics[width=\textwidth]{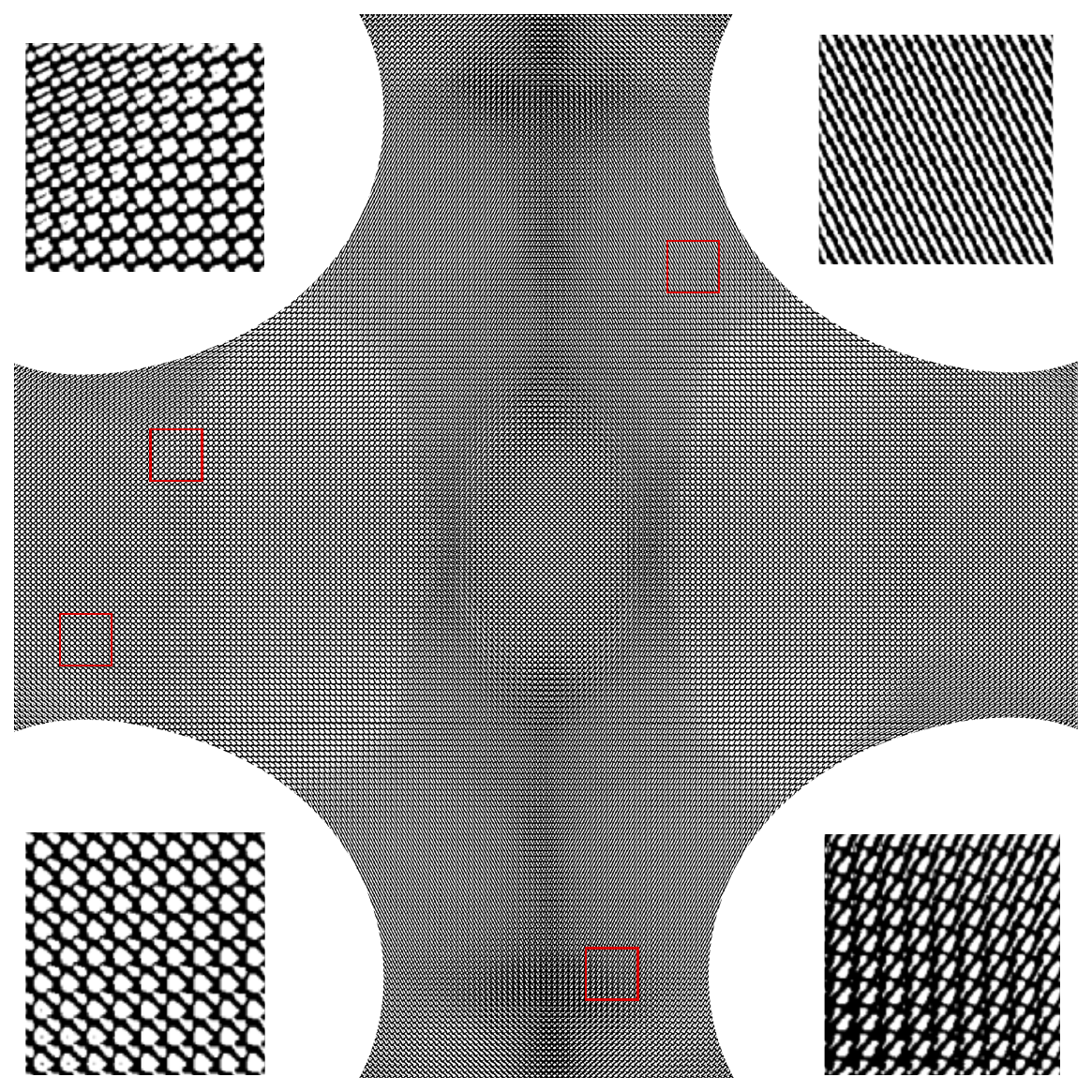}
\caption{}
\end{subfigure}

\caption{(a) We use a 1/4 mini-epoch setup, where one of four microstructure cells is optimized each epoch, cycling every four epochs. (b) The final result shows good cell connectivity without a major increase in runtime. (c) We then reduce the mini-epoch to 1/9 with good convergence still observed for the cantilever beam example. (d) We further reduce the mini-epoch to 1/16 for the second problem configuration, meaning the optimizer sees each microstructure cell once every 16 epochs. (e) With the 1/16 mini-epoch result, the gradation between adjacent microstructure is sufficiently small, allowing us to further upsample the microstructure without losing connectivity. This figure consists of a total of 36864 microstructure cells with a total of 58 million elements. Furthermore, with the smooth continuous spatial density function learned by the neural network, there is no theoretical upper limit to the sampling resolution. }

\label{fig:mm4}
\end{figure*}
\subsection{Multi-scale topology optimization}

\x{In this section, we delve into concurrent optimization of the two-scale structure. While the previous subsection emphasized the application of predefined stiffness tensors, our focus now shifts toward simultaneous optimization of the density distribution of the macro-scale structure and the micro-scale individual microstructure where similar optimization schemes are explored in \cite{xia2014concurrent,sivapuram2016simultaneous}. The neural network architecture remains the same as in the previous examples. We modify the loss function to include the compliance instead of the stiffness tensor target.   

The multiscale topology optimization problem considered here assigns a distinct microstructure to each macro finite element. Let the macrostructure $\Omega_M$ be discretized into $N_M$ elements, and for each macro element, we define a corresponding material microstructure $\Omega_m$ discretized into $N_m$ elements. The macro design variables are denoted by $\rho_M^i$ $(i = 1,2,\ldots,N_M)$, and the micro design variables (for each macro cell) by $\rho_m^j$ $(j = 1,2,\ldots,N_m)$. The optimization problem is stated as:

\begin{equation}
\begin{aligned}
&\text{Find: } \rho_M^i, \rho_m^j \quad (i = 1,2, \ldots, N_M; \, j = 1,2, \ldots, N_m) \\
&\text{Minimize: } c(\rho_M, \rho_m) \;=\; \frac{1}{2} \int_{\Omega_M} E_M(\rho_M, \rho_m)\,\varepsilon(\mathbf{u}_M)\,\varepsilon(\mathbf{u}_M)\,d\Omega_M \\
&\text{Subject to: } \\
&\quad a(\mathbf{u}_M, \boldsymbol{\nu}_M, E_M) \;=\; l(\boldsymbol{\nu}_M), 
\quad \forall\, \boldsymbol{\nu}_M \in H_{per}(\Omega_M, \mathbb{R}^d), \\
&\quad a(\mathbf{u}_m, \boldsymbol{\nu}_m, E_m) \;=\; l(\boldsymbol{\nu}_m), 
\quad \forall\, \boldsymbol{\nu}_m \in H_{per}(\Omega_m, \mathbb{R}^d), \\
&\quad V_M^*(\rho_M) \;=\; \int_{\Omega_M} \rho_M \, d\Omega_M \;-\; V_M \;\leq\; 0, \\
&\quad V_m^*(\rho_m) \;=\; \int_{\Omega_m} \rho_m \, d\Omega_m \;-\; V_m \;\leq\; 0, \\
&\quad 0 < \rho_M^{\min} \;\leq\; \rho_M^i \;\leq\; 1, 
\quad 0 < \rho_m^{\min} \;\leq\; \rho_m^j \;\leq\; 1.
\end{aligned}
\end{equation}

Here, $c(\rho_M,\rho_m)$ is the structural compliance. The macro topology is determined by $\rho_M$, and each macro element’s effective properties depend on its own micro topology described by $\rho_m$. $V_M^*(\rho_M)$ imposes a limit $V_M$ on the macro material usage, and $V_m^*(\rho_m)$ enforces a limit $V_m$ on the micro material consumption. The lower bounds $\rho_M^{\min}$ and $\rho_m^{\min}$ avoid numerical singularities. The displacement field at the macro scale is denoted by $\mathbf{u}_M$, with virtual displacement $\boldsymbol{\nu}_M \in H_{per}(\Omega_M, \mathbb{R}^d)$. The operator $a$ is a bilinear form representing the internal energy, and $l$ is a linear form associated with external forces. At the macro scale, the principle of virtual work leads to:

\begin{equation}
\begin{aligned}
\begin{cases}
  a(\mathbf{u}_M, \boldsymbol{\nu}_M, E_M) \;=\; 
      \displaystyle \int_{\Omega_M} E_M(\rho_M, \rho_m)\,\varepsilon(\mathbf{u}_M)\,\varepsilon(\boldsymbol{\nu}_M)\,d\Omega_M,\\[6pt]
  l(\boldsymbol{\nu}_M) \;=\; 
      \displaystyle \int_{\Omega_M} \mathbf{f}\,\boldsymbol{\nu}_M \, d\Omega_M \;+\; \int_{\Gamma_M} \mathbf{h}\,\boldsymbol{\nu}_M \, d\Gamma_M,\\[6pt]
  a(\mathbf{u}_m, \boldsymbol{\nu}_m, E_m) \;=\; 
      \displaystyle \int_{\Omega_m} E_m(\rho_m)\,\varepsilon(\mathbf{u}_m)\,\varepsilon(\boldsymbol{\nu}_m)\,d\Omega_m,\\[6pt]
  l(\boldsymbol{\nu}_m) \;=\; 
      \displaystyle \int_{\Omega_m} E_m(\rho_m)\,\varepsilon(\mathbf{u}_m^0)\,\varepsilon(\boldsymbol{\nu}_m)\,d\Omega_m.
\end{cases}
\end{aligned}
\end{equation}

In the above, $\mathbf{f}$ is the body force, and $\mathbf{h}$ is the boundary traction on the Neumann boundary $\Gamma_M$. The tensors $E_M(\rho_M,\rho_m)$ and $E_m(\rho_m)$ are derived via a material interpolation scheme akin to the modified SIMP approach, given by:

\begin{equation}
\begin{cases}
  E_M \;=\; \Big[c_0 + (\rho_M)^p\bigl(1 - c_0\bigr)\Big]\;E^H,\\[4pt]
  E_m \;=\; \Big[c_0 + (\rho_m)^p\bigl(1 - c_0\bigr)\Big]\;E_0,
\end{cases}
\end{equation}

where $E_0$ is the base material elasticity tensor, $c_0 = 10^{-9}$ prevents numerical singularity, and $p$ is the penalization exponent. The homogenized tensor $E^H$ depends on the local micro design:

\begin{equation}
\begin{aligned}
E^H &= \frac{1}{|\Omega_m|} \int_{\Omega_m} 
E_m(\rho_m)\;\bigl[\varepsilon(\mathbf{u}_m^0) \;-\; \\
&\quad \varepsilon(\mathbf{u}_m)\bigr]\;\bigl[\varepsilon(\mathbf{u}_m^0) \;-\; \varepsilon(\mathbf{u}_m)\bigr]\;d\Omega_m.
\end{aligned}
\end{equation}

A variety of gradient-based methods may be employed to solve this formulation. Thus, the sensitivities (first-order derivatives) of $c(\rho_M,\rho_m)$ and the volume constraints $V_M^*(\rho_M)$, $V_m^*(\rho_m)$ with respect to the design variables are needed. The partial derivatives of the objective and macro volume constraint with respect to $\rho_M$ read:

\begin{equation}
\frac{\partial c}{\partial \rho_M}
\;=\;
-\,\frac{1}{2}\;
\int_{\Omega_M} 
p\,(\rho_M)^{\,p-1}\,\bigl(1 - c_0\bigr)\,
E^H(\rho_m)\;\varepsilon(\mathbf{u}_M)\,\varepsilon(\mathbf{u}_M)\;
d\Omega_M,
\end{equation}

Likewise, the first-order derivatives of $c(\rho_M,\rho_m)$ and $V_m^*(\rho_m)$ with respect to the micro design $\rho_m$ are expressed as:

\begin{equation}
\frac{\partial c}{\partial \rho_m}-\frac{1}{2}\int_{\Omega_M}
\Bigl[c_0 + (\rho_M)^p\bigl(1 - c_0\bigr)\Bigr]
\frac{\partial\,E^H(\rho_m)}{\partial\,\rho_m}
\varepsilon(\mathbf{u}_M)\,\varepsilon(\mathbf{u}_M)
d\Omega_M,
\end{equation}

The derivative of the homogenized elastic tensor $E^H$ with respect to $\rho_m$ is given by:

\begin{equation}
\begin{aligned}
\frac{\partial E^H(\rho_m)}{\partial \rho_m}
\;=&\;
\frac{1}{|\Omega_m|}\,
\int_{\Omega_m}
p\,(\rho_m)^{\,p-1}\,\bigl(1 - c_0\bigr)\,
E_0\; \\
&\times \bigl[\varepsilon(\mathbf{u}_m^0) \;-\; \varepsilon(\mathbf{u}_m)\bigr]\,
\bigl[\varepsilon(\mathbf{u}_m^0) \;-\; \varepsilon(\mathbf{u}_m)\bigr]\,
d\Omega_m.
\end{aligned}
\end{equation}

With these sensitivities, standard gradient-based optimization procedures can be used to simultaneously update the macro design variables $\rho_M^i$ and each element’s corresponding micro design variables $\rho_m^j$. In implementation, the changes occur in implementing a multiscale FE solver, which computes the compliance $c$ for the entire structure and adds the volume fraction constraint $V^*_M$ for the macro scale structure. The combined loss function for concurrent multiscale topology optimization is:

\begin{equation}
\mathcal{L} = \frac{c}{c_{0}} + \alpha(\frac{V_i}{V^*_M}-1)^2+ \alpha(\frac{V_j}{V^*_m}-1)^2 + \beta\mathcal{L}_{bc,i}
\end{equation}

\noindent Furthermore, the macro-scale design is represented by a second neural network with the domain coordinate input reduced to 2D: $\textbf{X} = (x,y)$. While the frequency kernels $\textbf{K}$ and $\textbf{W}$ are reduced to align with the 2D input.

We demonstrate the multiscale topology optimization with three examples. The boundary condition is shown in Figure \ref{fig:ms} (a-c). The converged multi-scale structure is shown in Figure \ref{fig:ms} (d-f). One benefit of using a similar neural network to represent the macro scale design field is that it allows us to upsample as well. We use the upsampled macro scale density field as a filter to threshold the microstructures. Compared to filtering with the grid-based macro-scale design field, the continuous macro-scale density field allows smooth boundaries to be extracted that better follow microstructures. We observe that good cell compatibility is achieved in the multi-scale structure. 

Similar to the "mini-epoch" example in the previous subsection, we also demonstrate the neural network's interpolation ability with multi-scale topology optimization. We configure a 1/4 mini-epoch, which uses four microstructure cells per macro element. In Figure \ref{fig:mm4} (a), we choose one out of four microstructure cells every epoch to conduct homogenization and optimization. We cycle through each cell every four epochs. We run the same number of total epochs and the same boundary condition as the previous example. We observe that the converged result also consists of good cell connectivity with no significant increase in runtime. 

}

\begin{figure*}

\centering

\begin{subfigure}[t]{0.4\textwidth}
\centering
\includegraphics[width=\textwidth]{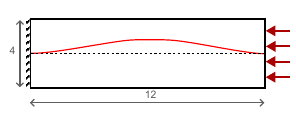}
\caption{}
\end{subfigure}
\qquad
\begin{subfigure}[t]{0.4\textwidth}
\centering
\includegraphics[width=\textwidth]{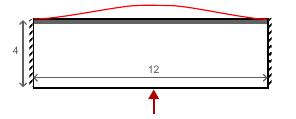}
\caption{}
\end{subfigure}

\begin{subfigure}[t]{0.4\textwidth}
\centering
\includegraphics[width=\textwidth]{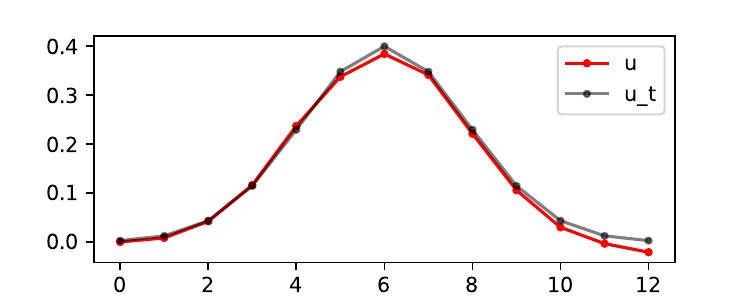}
\caption{}
\end{subfigure}
\qquad
\begin{subfigure}[t]{0.4\textwidth}
\centering
\includegraphics[width=\textwidth]{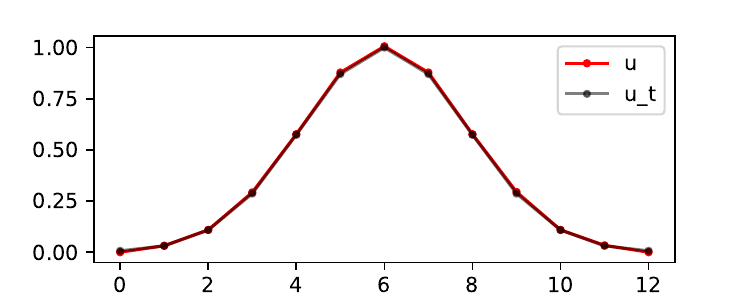}
\caption{}
\end{subfigure}

\begin{subfigure}[t]{0.4\textwidth}
\centering
\includegraphics[width=\textwidth]{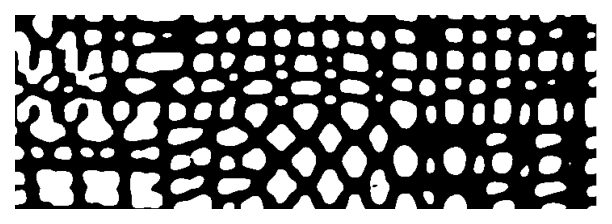}
\caption{}
\end{subfigure}
\qquad
\begin{subfigure}[t]{0.4\textwidth}
\centering
\includegraphics[width=\textwidth]{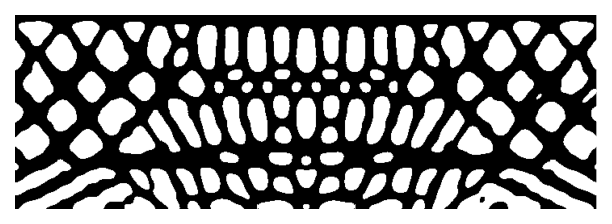}
\caption{}
\end{subfigure}

\begin{subfigure}[t]{0.4\textwidth}
\centering
\includegraphics[width=\textwidth]{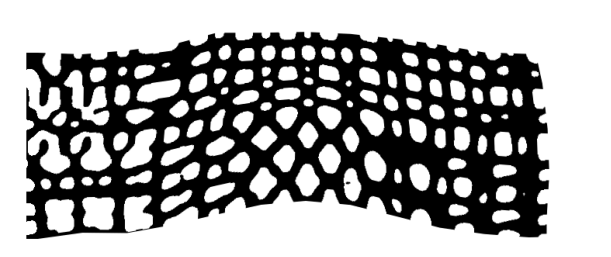}
\caption{}
\end{subfigure}
\qquad
\begin{subfigure}[t]{0.4\textwidth}
\centering
\includegraphics[width=\textwidth]{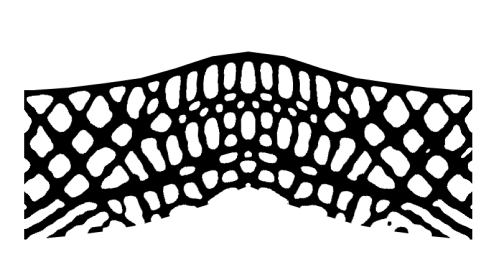}
\caption{}
\end{subfigure}

\caption{ (a)(b) Boundary conditions for two examples, with target displacement highlighted in red. (c)(d) Comparison of target and achieved displacement, showing minimal deviation, indicating accurate optimization. (e)(f) Converged metamaterial structures illustrating the optimized topology. (g)(h) Deformed metamaterial structures under applied loads, demonstrating the mechanical response of the optimized designs.}

\label{fig:mm}
\end{figure*}

\subsection{Metamaterial system design}
\x{
Metamaterials are engineered materials whose unique properties stem from the arrangement of their microstructures rather than the materials they are made of \cite{zheludev2010road}. Mechanical metamaterials are a specific type designed to achieve mechanical characteristics that cannot be attained through their base material alone. Prior work by Wang et al. demonstrated metamaterial design of metamaterials to match a target displacement through the microstructure selection with a deep generative model \cite{wang2020deep}. We are motivated by prior work in designing metamaterials to match a target displacement with our neural network. For the metamaterial design problem, similar to the prior work by Wang et al., we limit the scope to microstructure design. One difference is that since we are designing the microstructure real-time, we also need an additional optimization objective of bulk modulus maximization to ensure the result is overall structurally sound. Otherwise, with only the displacement target as the objective, the result may lack structural performance in other loading conditions. 

The objective function first consists of a target displacement $\textbf{u}_t$ where the displacement $\textbf{u}$ of the multiscale structure is trying to match. Same with Wang et al., we minimize the difference in displacement with an L2 loss:
\begin{equation}
\mathcal{F} = ||\mathbf{\gamma}\circ\mathbf{u}-\mathbf{u}_t||^2_2
\end{equation}

\noindent Following Wang et al., we only design the microstructure in metamaterial system design. Thus we modify their sensitivity function to compute the derivative of the objective function w.r.t micro design field $x_j$:

\begin{equation}
\begin{aligned}
\frac{\partial \mathcal{F}}{\partial x_j}
  &= -2 \bigl[\gamma \odot (u - u_t)\bigr]^T 
        K^{-1} \frac{\partial K}{\partial x_j} \,u,\\
&\quad \text{where}\quad 
  \frac{\partial K_e}{\partial x_j} 
    = B^T \frac{\partial C_e}{\partial x_j} \,B.
\end{aligned}
\end{equation}

The combined loss function also resembles the previous examples of bulk modulus minimization, but it includes an additional term for displacement mismatch. 

\begin{equation}
\mathcal{L} = \frac{1}{n}\sum^n \frac{c_i}{c_{0,i}} + \alpha||\mathbf{\gamma}\circ\mathbf{u}-\mathbf{u}_t||^2_2 + \alpha(\frac{V_i}{V^*_i}-1)^2 + \beta\mathcal{L}_{bc,i}
\end{equation}

\noindent Specifically for displacement matching, $\mathbf{\gamma}$ is a vector with the value one at the degrees of freedom corresponding to the displacement targets and with zeros everywhere else. 

We configure two examples with different boundary conditions Figure \ref{fig:mm} (a)(b). The target displacement is highlighted in red in each boundary condition figure. In the first example, the boundary condition and displacement target are similar to one of the examples in Wang et al. \cite{wang2020deep}. A load is applied on the right-hand side with a target displacement along the center of the structure. We configure the second example with load applied to the bottom center and target displacement prescribed to the top surface with a region of enforced solid highlighted in gray. Some potential applications for the second boundary condition include flexible haptics devices and jet engine intake ramps. We plot out the target and achieved displacement in Figure \ref{fig:mm} (c)(d), where we observe minimal errors. The converge metamaterial is shown in Figure \ref{fig:mm} (e)(f) with deformed metamaterial shown in Figure \ref{fig:mm} (g)(h). 
}
\section{LIMITATIONS AND FUTURE WORK}

In this study, our primary emphasis lies in showcasing our capability to generate interconnected microstructures amidst demanding design constraints. The convergence of the multi-scale structure can be further explored. In a prior work \cite{chen2023cond}, the authors explored using a conditioning field to speed up convergence on single-scale topology optimization problems. A conditioning field can also be used to apply to multi-scale structures. One possible conditioning field is the rank-N laminates. The neural network can help connect the Rank-N laminates. In another prior work, \cite{joglekar2023dmftonn}, Physics Informed Neural Network (PINN) is used to directly optimize topology. It is worth exploring if PINN can also be used in multi-scale optimization, whether in macro-scale designing stiffness tensors or in micro-scale designing microstructures. 

Manufacturing constraints can also be explored. In prior works \cite{chen2023concurrent}, the neural network-based topology optimization allows the computation of density gradients; therefore, overhang control can be applied for additive manufacturing. Overhang control can also be applied to microstructures such that each individual cell can be additively manufactured without suffering from local print failures. Multi-material and multi-physics can also be integrated into the proposed approach.

\section{CONCLUSIONS}

In conclusion, our study introduces a novel approach to direct multi-scale topology optimization with neural networks. Through the utilization of neural networks, particularly in inverse homogenization tasks, we have effectively enhanced microstructure compatibility within challenging design conditions. Our methodology involves training a topology neural network on a per-case basis, enabling the representation of diverse microstructure designs across the design domain. We demonstrated the efficacy of the approach with examples. Notably, our approach enables the optimization of individual microstructure cells based on specified elasticity tensors while accommodating rotation, thereby facilitating a high degree of design flexibility.

\section*{Acknowledgment}
This research was funded by Lockheed Martin Corporation MRA19001RPS004. 

We want to thank Professor Liwei Wang for insightful discussions and assistance in multi-scale topology optimization. 

\bibliography{sn-bibliography}

\end{document}